%% file: main.tex
\documentclass{article} 
\usepackage{iclr2023_conference,times}

\input{math_commands.tex}

\usepackage{hyperref}
\usepackage{url}

\usepackage{wrapfig}
\usepackage{standalone}
\usepackage{algorithm}
\usepackage{algpseudocode}
\usepackage{listings}
\usepackage{tikz}

\usepackage{subcaption}

\usepackage{booktabs}
\usepackage{multirow}

\usepackage[inline,shortlabels]{enumitem}

\usetikzlibrary{snakes,arrows,shapes}
\usetikzlibrary{chains, scopes}
\usetikzlibrary{backgrounds}
\usetikzlibrary{shapes,arrows}

\definecolor{graphnodecolor}{HTML}{336699}

\definecolor{codegreen}{rgb}{0,0.6,0}
\definecolor{codegray}{rgb}{0.5,0.5,0.5}
\definecolor{codepurple}{rgb}{0.58,0,0.82}
\definecolor{backcolour}{rgb}{0.95,0.95,0.92}

\makeatletter
\newcommand{\srcsize}{\@setfontsize{\srcsize}{4pt}{4pt}}
\makeatother

\lstdefinestyle{mystyle}{
    backgroundcolor=\color{backcolour},
    commentstyle=\color{codegreen},
    keywordstyle=\color{magenta},
    numberstyle=\tiny\color{codegray},
    stringstyle=\color{codepurple},
    basicstyle=\ttfamily\footnotesize,
    captionpos=b,
    keepspaces=true,
    numbers=left,
    numbersep=5pt,
    frame=single
}
\lstdefinestyle{modelstyle}{
    backgroundcolor=\color{backcolour},
    commentstyle=\color{codegreen},
    keywordstyle=\color{magenta},
    stringstyle=\color{codepurple},
    basicstyle=\ttfamily\footnotesize,
    captionpos=b,
    numbers=none,
    keepspaces=true,
    frame=single
}
\title{Discovering Graph Generation Algorithms}

\author{Mihai Babiac, Karolis Martinkus \& Roger Wattenhofer\\
ETH Zurich \\
\texttt{\{mbabiac,martinkus,wattenhofer\}@ethz.ch}
}

\iclrfinalcopy 
\begin{document}

\maketitle

\begin{abstract}
We provide a novel approach to construct generative models for graphs. Instead of using the traditional probabilistic models or deep generative models, we propose to instead find an algorithm that generates the data. We achieve this using evolutionary search and a powerful fitness function, implemented by a randomly initialized graph neural network. This brings certain advantages over current deep generative models, for instance, a higher potential for out-of-training-distribution generalization and direct interpretability, as the final graph generative process is expressed as a Python function. We show that this approach can be competitive with deep generative models and under some circumstances can even find the true graph generative process, and as such perfectly generalize.
\end{abstract}

\section{Introduction}
Generating new samples of graphs similar to a given set of graphs is a long-standing problem, which initially was tackled with various statistical models, such as the Erdős/Rényi model \citep{erdos59a, hollandStochasticBlockmodelsFirst1983, eldridgeGraphonsMergeons2017}. While such models lend themselves well to formal analysis, they do not closely fit real-world graph distributions. More recently, deep generative models have proven to fit graph distributions well \citep{youGraphRNNGeneratingRealistic2018, liaoEfficientGraphGeneration2020,simonovskyGraphVAEGenerationSmall2018a,martinkusSPECTRESpectralConditioning2022,haefeli2022diffusion,vignac2022digress}. However, similar to other deep models, they are not interpretable and struggle to generalize to graph sizes outside of the training distribution.

\begin{wrapfigure}{r}{0.3525\textwidth}
\hfill
\begin{subfigure}{0.325\textwidth}
\begin{lstlisting}[language=Python, basicstyle=\ttfamily\tiny]
def main():
  for i in range(N):
    int04 = i % W
    bool00 = W != 0
    call()

def call():
    for j in range(i):
        int05 = i - j
        if bool01:
            pass
        bool01 = int05 == W
        if bool01:
            int06 = i
            add_edge(j, int06)
        else:
            bool01 = int04 != 0
            if bool01:
                add_edge(i, j)
\end{lstlisting}
\end{subfigure}
\caption{An example program from the search space that
generates a grid graph. \texttt{i, j, N} and \texttt{W} are aliases for
\texttt{int00-03}, where \texttt{N} is the number of nodes in the grid and
\texttt{W} is the width of the grid. Lines 1, 2, 7, and 8 are hard-coded, while
the contents of the two for loops are what we are searching over.}
\label{lst:search_space_sample}
\vspace{-1.0em}
\end{wrapfigure}

In this work, we propose an alternative approach. Given a set of graphs, we aim to learn a single algorithm, represented as a Python function, that can generate that given set of graphs as well as new examples. If we find an algorithm that fits the data well we can directly inspect it and potentially learn about the generative process that originally created the data. This can be quite useful for example in social sciences. Additionally, the produced algorithm will have a predictable behavior if the input parameters go beyond those observed during training.

We achieve this by factorizing the graph construction algorithm into two loops (Figure~\ref{lst:search_space_sample}) over nodes and their potential neighbors. Evolutionary search is used to assemble the logic inside the loops to define when graph edges are added or removed. This setup is closely related to genetic programming \citep{sobaniaComprehensiveSurveyProgram2022} where programs are evolved to match input-output examples, especially the grammar-guided \citep{whigham1995grammatically, forstenlechnerGrammarDesignDerivation2016, fentonPonyGE2GrammaticalEvolution2017} and linear \citep{lalejiniTagaccessedMemoryGenetic2019, hernandezRandomSubsamplingImproves2019, dolsonExploringGeneticProgramming2019, fergusonCharacterizingEffectsRandom2020} genetic programming approaches. In contrast to stack-based approaches \citep{perkis1994stack, spectorPush3ExecutionStack2005} they aim to generate code in a standard programming language such as Python. In general, evolutionary search has proven to be a powerful method for code search. It can even discover neural network architectures and their training functions \citep{realAutoMLZeroEvolvingMachine2020}. Our work is also related to methods that aim to learn a program that draws a figure \citep{ellis2018learning, ellisDreamCoderGrowingGeneralizable2020}. However, we want a program that captures a family of graphs, not just a single figure.

Note that there are some genetic programming packages that allow generating Python code
\citep{fentonPonyGE2GrammaticalEvolution2017}.
However, these proved to be too slow and not well suited for our problem.

We implement our method as a highly-optimized C++ framework and experimentally validate it on synthetic and real-world datasets. When applied to grid graphs with known grid sizes, our search finds algorithms that generalize perfectly to arbitrary grid sizes.

\section{Method}
In this section, we explain our representation, and how the evolutionary search is performed.

\subsection{Algorithm Representation}
\begin{figure}
\centering
    \begin{subfigure}{0.3\textwidth}
    \centering
        \begin{lstlisting}[language=Python, basicstyle=\ttfamily\tiny,gobble=12]
            int05 = int00 - int01
            if bool01:
                pass
            bool01 = int05 == int03
            if bool01:
                int06 = int00
                add_edge(int01, int06)
            else:
                bool01 = int04 != 0
                if bool01:
                    add_edge(int00, int01)
                int05 = 4
        \end{lstlisting}
        \caption{Python code}
        \label{fig:transition_orig_code}
    \end{subfigure}
    \hspace{0.5em}
    \begin{subfigure}{0.3\textwidth}
    \centering
        \scalebox{0.35}{
        \begin{tikzpicture}[node distance=3mm,
                            every node/.style={draw, minimum height=2em, minimum width=3cm},
                            every join/.style=->,
                            >=latex]
            { [start chain=1 going below]
                \node [on chain, join] (n0) {\texttt{EmptyNode}};
                \node [on chain, join] (n1) {\texttt{StatementNode}};
                \node [on chain, join] (n2) {\texttt{IfNode}};
                \node [on chain, join] (n3) {\texttt{StatementNode}};
                \node [on chain, join] (n4) {\texttt{IfNode}};
            }

            { [start chain=2 going below]
                \node [on chain=2, join, anchor=north west, below right=3mm and 7mm of n4.south] (n5) {\texttt{StatementNode}};
                \node [on chain=2, join] (n6) {\texttt{StatementNode}};
            }
            \draw [->, rounded corners=5pt] (n4) |- (n5.west);

            { [start chain=3 going below]
                \node [on chain=3, join, anchor=north, below=6mm of n6.south] (n7) {\texttt{StatementNode}};
                \node [on chain=3, join] (n9) {\texttt{IfNode}};
            }
            \draw [->, rounded corners=5pt] (n4) |- (n7.west);

            \node [anchor=north west, below right=3mm and 7mm of n9.south] (n10) {\texttt{StatementNode}};
            \draw [->, rounded corners=5pt] (n9) |- (n10.west);

            { [continue chain=3 going below]
                \node [on chain=3, join, anchor=north west, below left=3mm and 7mm of n10.south] (n11) {\texttt{StatementNode}};
            }
        \end{tikzpicture}
        }
        \caption{Code tree}
        \label{fig:transition_code_tree}
    \end{subfigure}
    \hspace{0.8em}
    \begin{subfigure}{0.3\textwidth}
    \centering
        \begin{lstlisting}[language=Python, basicstyle=\ttfamily\tiny,gobble=12,firstnumber=0]
            MINUS_I(5, 0, 1)
            JUMP_ABS(2)
            EQ_I(1, 5, 3)
            JUMPZ(1, 7)
            ASGN_I(6, 0)
            ADD_EDGE(1, 6)
            JUMP_ABS(13)
            NEQ_IMM_I(1, 4, 0)
            JUMPZ(1, 11)
            ADD_EDGE(0, 1)
            JUMP_ABS(11)
            CONST_I(5, 4)
            JUMP_ABS(13)
        \end{lstlisting}
        \caption{Compiled instructions}
        \label{fig:transition_compiled}
    \end{subfigure}
    \caption{The internal representations of a program. Left: Python code. Middle: internal representation as code tree. Right: compiled instructions used for running the code tree.}
    \label{fig:transition}
\end{figure}
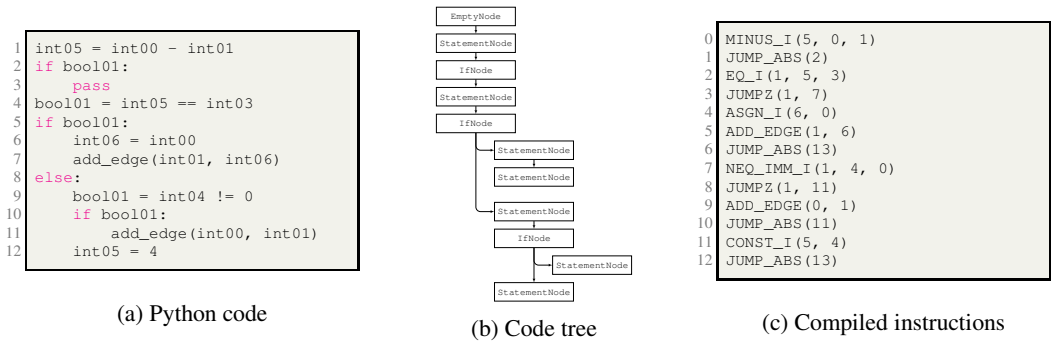

Ideally, the representation should be fast to execute, easy to mutate, and human-readable. Unfortunately, some of these goals can be hard to align. Instead, we choose to build a representation that is easy to mutate and can be easily converted to either a human-readable or an efficiently executable format. This representation consists of a tree of nodes (Figure~\ref{fig:transition_code_tree}), in which each node corresponds (with few exceptions) to one line of code in a Python program (Figure~\ref{fig:transition_orig_code}), where the lines contain either simple operations or if-statements. The structure of the tree matches the logical structure of the code that is implied through indentation: the tree branches out at if-statements and is chain-like otherwise. This allows us to execute the lines of code in the correct sequence by simply correctly traversing the tree (Appendix~\ref{appx:exec_code}).
As shown in Figure~\ref{lst:search_space_sample}, our algorithms consist of two for-loops, which can be nested. The code tree can represent the contents of such a for-loop, but the loop itself is hard-coded in the function that executes the individual.

To respect the code's branching behavior, we define three types of nodes: statement nodes, if-nodes, and empty/root nodes. Each node has a fixed number of named slots for its children, where each slot contains either a pointer to the child node or a marker, the \texttt{NULL} pointer if the child is missing.

A \textit{statement-node} contains an instruction to be executed. They have a single child slot, named \texttt{nextInBranch}, which points to the next line of code in the current branch (i.e. at the current level of indent for Python code). \textit{Root-nodes} are similar to statement-nodes but do not contain any instructions. They are at the root of a code tree, and their purpose is to simplify inserting new code at the beginning of the program. \textit{If-nodes} contain the address of a boolean condition variable, which decides which branch to take, and they have three child slots. The slots correspond to the ``then'' branch, to the ``else'' branch, and to the next line of code at the same level of indent and are named \texttt{thenBranch}, \texttt{elseBranch} and \texttt{nextInBranch} respectively.

Given this structure, we convert the code to a Python-like representation as follows: We traverse the tree in the order \texttt{currentNode} $\to$ \texttt{thenBranch} $\to$ \texttt{elseBranch} $\to$ \texttt{nextInBranch} and, keeping track of the current nesting level, we simply output the associated lines of code. 

Similarly to conversion into Python, it is possible to traverse the tree and directly execute it. However, this has some drawbacks and can be slow. Instead, we propose to use a simple compilation step (Figure~\ref{fig:transition_compiled}) to improve evaluation speed inside the for-loops and thus the search speed. We discuss this and other representation implementation details in Appendix~\ref{appx:repr_and_exec}.

The instructions allowed in our representation are listed in Appendix~\ref{appx:instructions}. They cover the usual arithmetic, relational and Boolean operations, if-else statements, and function calls for generating random numbers.
We restrict the available variables to a limited set that is strongly typed. Note that in our setup all variables are global. For constructing the graph, we use an adjacency matrix which the algorithm can manipulate and interrogate through dedicated functions. Notably, we disallow while-loops, additional for-loops (except the two hard-coded ones), and jumps to ensure all algorithms have finite running time.


\subsection{Evolutionary Search}

The evolutionary search is started by initializing each individual as an empty algorithm. 
Every round we use tournament selection based on the graph fitness. The tournaments are stochastic, where softmax is applied on the fitness values and the winners are then randomly sampled. Taking inspiration from simulated annealing \citep{kirkpatrickOptimizationSimulatedAnnealing1983} we gradually decay the softmax temperature during the search to ensure better exploration during the initial phases. The children are generated only through mutation of the parents, without crossovers, as done by \citet{realAutoMLZeroEvolvingMachine2020}. The mutations are effectively point-wise. We either delete, insert or change a command by changing one of its variables. In each round, the whole generation is replaced with the children, with some number of the best (elite) individuals being preserved. More information is provided in Appendix~\ref{appx:evo}.

To determine the fitness, we use the individual to generate a set of graphs and compare those to the training set graphs. 
Determining if two graph distributions are the same is a hard task \citep{obrayEvaluationMetricsGraph2022}. For our fitness function, we chose to use a randomly initialized GIN graph neural network \citep{xuHowPowerfulAre2019} as a powerful feature extractor \citep{thompsonEvaluationMetricsGraph2022} to embed the training graphs and graphs produced by the generated code and then compare the two resulting distributions (Appendix~\ref{appx:fitness}).
The fitness value is computed by using a Radial basis function (RBF) kernel and computing Maximum Mean Discrepancy (MMD) between the two embedding distributions. 
The use of a randomly initialized model helps us avoid cumbersome training of the graph neural network in an adversarial setting, but still provides us with a well-encompassing feature extractor, that can implicitly capture various graph structures and even node and edge features. This contrasts with the alternative of manually computing various pre-defined graph metrics, such as clustering coefficient and node degree distributions.
Note, that \citet{dziugaite2015training} have shown that it is possible to train generative adversarial networks for image generation by relying only on MMD instead of a trained discriminator. So MMD can be a well-encompassing metric for generative model training.

\section{Experiments}
Our work has similarities to autoregressive deep graph generative models \citep{liaoEfficientGraphGeneration2020,youGraphRNNGeneratingRealistic2018}, as our search is biased towards iterative algorithms. Thus, we use the experimental setup of GRAN \citep{liaoEfficientGraphGeneration2020} and compare it to GRAN and GraphRNN \citep{youGraphRNNGeneratingRealistic2018} as baselines for our experiments. 
The datasets and their 80/20 split between training and test data match the ones used by GRAN \citet{liaoEfficientGraphGeneration2020}. 
We only adjust the Grid dataset to filter out isomorphic copies of grids from the dataset to avoid data leakage between the training and test sets. To make training and search faster we consider grids with 9 to 81 nodes. We re-train the baselines on this modified dataset using the original hyperparameters.
In our case, we do not make use of a validation set, as in our setup it is redundant, since we do not expect overfitting. During the search procedure, similarly to neural network training, we follow a mini-batch approach, where in every round of search a random subset of 16 test graphs is used to compute the fitness scores. This helps to avoid performing expensive MMD computation over the whole training set at once. See Appendix~\ref{appx:hyperparams} for other search hyperparameters. The code is publicly available\footnote{\url{https://github.com/MihaiBabiac/graph_gen_algo}}.

Table~\ref{tab:result_comparison} shows that our discovered graph algorithms perform quite well on the simpler grid and lobster graph datasets. While the deep learning models achieve slightly better statistical similarity, the discovered algorithms are competitive and respect constraints, such as graphs having no triangles (clustering coefficient of $0$). Importantly, they do all this, while providing an interpretable graph generator. However, the approach struggled with more complex protein graphs. This might indicate insufficient exploration under some circumstances.

\begin{table}[tpb]
    \centering
    \begingroup
    \resizebox{\linewidth}{!}{
        \setlength{\tabcolsep}{0.5\tabcolsep}
        \begin{tabular}{l *{12}{c}}\toprule
                     &                    \multicolumn{4}{c}{Grid}                                                       &                        \multicolumn{4}{c}{Lobster}                                       &            \multicolumn{4}{c}{Protein}              \\
            \cmidrule(lr){2-5} \cmidrule(lr){6-9} \cmidrule(lr){10-13}
                     &          Deg.          &          Clus.          &       Orbit            &          Spec.         &          Deg.          &        Clus.    &        Orbit           &       Spec.           &          Deg.          &        Clus.          & Orbit         &        Spec.           \\
            \midrule
            GraphRNN &  {$\mathbf{5.34e^{-4}}$}          &  {$\mathbf{0.0}$}           &  $\mathbf{8.41e^{-5}}$ &   $\mathbf{3.93e^{-2}}$&  $\mathbf{9.26e^{-5}}$ & $\mathbf{0.0}$  & $\mathbf{2.19e^{-5}}$  & $\mathbf{1.14e^{-2}}$ &  $1.06e^{-2}$          & $0.14      $          &  0.88         & $1.88e^{-2} $          \\
            GRAN     &  ${2.42e^{-2}}$ &  $\mathbf{0.0}$           &  $4.51e^{-2}$          &   $6.87e^{-2}$         &  $3.73e^{-2}$          & $\mathbf{0.0}$  & $7.67e^{-4}$           & $2.71e^{-2}$          &  $\mathbf{1.98e^{-3}}$ & $\mathbf{4.86e^{-2}}$ &  \textbf{0.13}& $\mathbf{5.13e^{-3}} $ \\
            Ours     &  $5.10e^{-3}$          &  $\mathbf{0.0}$         &  $3.38e^{-3}$          &   $0.1       $         &  $2.77e^{-3}$          & $\mathbf{0.0}$  & $2.31e^{-2}$           & $3.41e^{-2}$          &  $0.42      $          & $1.07      $          &  1.14         & $0.31       $          \\
            \bottomrule
        \end{tabular}
    }
    \endgroup
    \caption{Comparison with deep graph generative models. The results for the baseline models are taken from \citet{liaoEfficientGraphGeneration2020}.}
    \label{tab:result_comparison}
\end{table}
\begin{figure}[tpb]
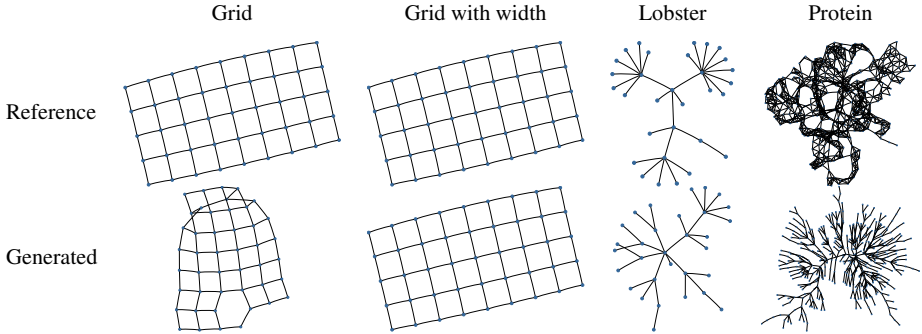

    \centering
    \resizebox{0.9\textwidth}{!}{
    \begin{tabular}{lcccc}
        {} & {Grid} & {Grid with width} & {Lobster} & {Protein}  \\
        {} & {} & {} & {} & {} \\[-0.5em]
         Reference &
         $\vcenter{\hbox{\includestandalone[height=2.3cm]{figures/true_Grids}}}$ &
         $\vcenter{\hbox{\includestandalone[height=2.3cm]{figures/true_Grids_with_info}}}$ &
         $\vcenter{\hbox{\includestandalone[height=2.3cm]{figures/true_Lobsters}}}$ &
         $\vcenter{\hbox{\includestandalone[height=2.3cm]{figures/true_Proteins}}}$
         \\
         Generated &
         $\vcenter{\hbox{\includestandalone[height=2.3cm]{figures/fake_Grids}}}$ &
         $\vcenter{\hbox{\includestandalone[height=2.3cm]{figures/fake_Grids_with_info}}}$ &
         $\vcenter{\hbox{\includestandalone[height=2.3cm]{figures/fake_Lobsters}}}$ &
         $\vcenter{\hbox{\includestandalone[height=2.3cm]{figures/fake_Proteins}}}$
         \\

    \end{tabular}}
    \caption{Comparison of reference graphs from the test set to graphs generated by algorithms found with our method. Graphs in the same column have the same number of nodes.}
  \label{fig:graph_comparison}
\end{figure}

\begin{wrapfigure}{r}{0.28\textwidth}
\hfill
\vspace{1em}
\begin{subfigure}{0.275\textwidth}
\begin{lstlisting}[language=Python, basicstyle=\ttfamily\tiny]
def main():
  for i in range(N):
    remove_edge(int06, int10)
    add_edge(i, int06)
    int07 = i % W
    int09 = i - W
    add_edge(int09, i)
    int10 = int07 + int06
    int06 = i + int05
\end{lstlisting}
\end{subfigure}
\caption{The discovered algorithm for grids, when the grid width $W$ is supplied alongside the number of nodes $N$.}
\label{lst:grid_algo}
\end{wrapfigure}
Even though we are able to find an algorithm that produces statistically similar graphs to grids, when supplied just with node count, the graphs are not true grids (Figure~\ref{fig:graph_comparison}). In our setup, the algorithm can also be conditioned on additional input values. If we instead perform our search over algorithms that take both the node count and the grid width as inputs, our method manages to recover the true generative algorithm (Figure~\ref{fig:alg_grids}). While it was discovered only using a dataset with up to 81 nodes, it can generalize to produce perfect grids of any size, showcasing the potential benefit of using program synthesis for the graph generation. This is impossible with current deep generative models, which are also incapable of generating perfect, rectangular grids even without considering extrapolation \citep{liaoEfficientGraphGeneration2020}.
This suggests that providing extra inputs to the algorithm can provide a user with additional control over the discovered function's outputs, and also makes the search simpler.
The discovered algorithm in Figure~\ref{fig:alg_grids} is also quite interpretable. This is the case for most of the algorithms we have discovered (Appendix~\ref{appx:discovered_algos}). 

\section{Conclusion}
In this work, we show that it is possible to use program synthesis to discover interpretable graph generation algorithms. In certain cases even the true generative processes can be discovered, resulting in ideal extrapolation.
The performance on more complex graph families can likely be improved, by more carefully tuning the fitness function to make it smoother. As shown by \citet{obrayEvaluationMetricsGraph2022} MMD metrics can be quite sensitive to their hyperparameters. Another strong extension could be a compilation of a library of primitive functions that are useful for graph construction. For example, if we provided a function to factorize a number, discovering an algorithm for grid generation when no additional inputs are given would likely be as easy as when the grid width is provided.
In principle, the approach can also be applied to attributed graph generation, which is left for future work.

\bibliography{our}
\bibliographystyle{iclr2023_conference}

\clearpage
\newpage
\appendix

\section{Details on Algorithm Representation and Execution}
\label{appx:repr_and_exec}
Similarly to \citet{realAutoMLZeroEvolvingMachine2020} we implement our search procedure in C++ to ensure efficiency. As our chosen representation and implementation are tailored to our problem setting of finding graph generation algorithms it is easier 
to ensure that our search procedure is efficient and satisfies our goals of fast selection, mutation, and execution procedures, as well as a human-readable output format.
Our implementation will be made publicly available.

\subsection{Efficient execution}
\label{appx:exec_code}
As noted in the main text, we can traverse the tree and run it using the naive algorithm in Listing~\ref{lst:code_tree_traversal}. While this algorithm functions correctly, it has some drawbacks. Note that the traversal presented here uses recursion, which can have significant overhead, especially when the amount of useful processing done per call is small. We can solve this by using an explicit stack (instead of the implicit call stack) to keep track of which nodes should be visited next. In practice, we have noticed that the traversal-based execution described above has suboptimal performance, even after making the stack explicit and even by reusing the same stack across runs in order to avoid memory allocations. The likely cause of the slowdown is the layout of the nodes in memory: the nodes are not at sequential memory locations, so the traversal will not use the CPU cache optimally.
\begin{figure}[htb]
    \begin{lstlisting}[language=Python]
        def run_code_tree(node, memory):
            if node is None:
                return
    
            if type(node) is IfNode:
                if memory.bools[node.conditionVar]:
                    run_code_tree(node.thenBranch, memory)
                else:
                    run_code_tree(node.elseBranch, memory)
            elif type(node) is StatementNode:
                exec_instruction(node.instruction, memory)
    
            # For all node types go to nextInBranch
            run_code_tree(node.nextInBranch, memory)
    \end{lstlisting}
    \caption{Executing a code tree through recursive traversal (pseudocode). \texttt{memory} contains the current execution state.}
    \label{lst:code_tree_traversal}
\end{figure}

Our chosen solution is to perform a ``compilation'' step, converting the code tree to an array of instructions (bytecode) which can then be efficiently executed. Of course, compiling the tree incurs an overhead, but it is offset by more quickly executing the code tree as part of a for-loop.

The idea behind the compilation step is to use jump instructions to convert the if-statements into linear code. Whenever we encounter an if-node with a \texttt{thenBranch} or \texttt{elseBranch}, we insert a conditional jump that skips to the appropriate location in the program. At the same time, we store the node that comes after the if statement in a stack so that we know where to jump once the execution of a branch has finished. The complete algorithm is provided in Listing~\ref{lst:compiling}.

Figure~\ref{fig:transition} illustrates how a piece of Python code is represented as a code tree and how this can then be converted to a linear sequence of instructions. The Python code is similar to that in the \texttt{call} function from Figure~\ref{lst:search_space_sample}, but the variables' internal names are used, and we have added another statement on line~12. We made this last modification to highlight that the compilation algorithm inserts redundant instructions, such as the jump at location 10. It is practically a no-op since the program flow would continue to that instruction by itself. It would have been possible to eliminate such instructions, but since they only appear at the end of branches (which will likely not make up a significant proportion of the code), we decided not to pursue such an optimization.

\begin{figure}[htb]
    \hfill
    \begin{lstlisting}[language=Python]
        def compile_code_tree(tree)
            jump_target_stack = [None]
            jump_list = []
            program = []
    
            def add_jump(jump_op, target_node, cond_var=None):
                jump_list.append((len(program), target_node))
                if cond_var is None:
                    program.append([jump_op, None])
                else:
                    program.append([jump_op, cond_var, None])
    
            for node in tree:
                node.instrIdx = len(program)
                if type(node) is IfNode:
                    if node.nextInBranch:
                        next_node = node.nextInBranch
                    else:
                        next_node = jump_target_stack.pop()
    
                    if node.thenBranch and node.elseBranch:
                        jump_target_stack.extend([next_node]*2)
                        add_jump("JUMPZ", node.elseBranch,
                                 node.conditionVar)
                    elif node.thenBranch:
                        jump_target_stack.append(next_node)
                        add_jump("JUMPZ", next_node,
                                 node.conditionVar)
                    elif node.elseBranch:
                        jump_target_stack.append(next_node)
                        add_jump("JUMPNZ", next_node,
                                 node.conditionVar)
                    else:
                        add_jump("JUMP_ABS", next_node)
                elif type(node) is StatementNode:
                    program.append(node.instruction)
                    if not node.nextInBranch:
                        next_node = jump_target_stack.pop()
                        add_jump("JUMP_ABS", next_node)
    
            for jump_instr_idx, target in jump_list:
                if target is None:
                    target_instr_idx = len(program)
                else:
                    target_instr_idx = target.instrIdx
    
                program[jump_instr_idx][-1] = target_instr_idx
    
            return program
    \end{lstlisting}
    \caption{Python pseudocode for compiling a code tree. For simplicity, we denote instructions as lists of one op-code and operands. Assumes a code tree has an iterator for traversing it in the order \texttt{currentNode} $\to$ \texttt{thenBranch} $\to$ \texttt{elseBranch} $\to$ \texttt{nextInBranch}.}
    \label{lst:compiling}
\end{figure}
\subsection{Execution environment and instructions}
\label{appx:instructions}
So far, we have looked at how to compile a code tree into a sequence of instructions but skimmed over the details of the instructions and the execution environment (the memory). This subsection fills those gaps.

To make the concepts clear, it is best to start with an overview. We represent the memory as an object with different fields, where each field contains a different type of data. The instructions are similar to the machine instructions of a CPU and consist of an operation (op-code) and operands. Depending on the op-code, the operands can be interpreted as addresses inside one of the fields or as constants. Moreover, addresses refer to specific fields in the memory depending on the op-code. For example, \texttt{LT\_IMM\_F(5, 3, 0.125)} is the representation of \texttt{bool05 = float03 < 0.125}, while \texttt{PLUS\_I(1, 2, 3)} corresponds to \texttt{int01 = int02 + int03}, where \texttt{bool}, \texttt{int} and \texttt{float} indicate different memory fields.

\paragraph{Memory.}
In detail, the memory consists of:
\begin{enumerate*}[(1)]
    \item a program counter;
    \item three sets of ``registers'' (i.e. arrays), one for integers, booleans, and floating-point numbers, respectively;
    \item the state of a pseudo-random number generator (PRNG);
    \item per-node storage in the form of an array of integers;
    \item the representation of the graph being built;
\end{enumerate*}

The \textit{program counter} is the index of the instruction being executed as part of the compiled program. It is incremented by one automatically to progress the flow of the program but can be manipulated by jump instructions to implement branching.

The \textit{registers} act as the main inputs and outputs for the instructions, and their size can be configured. These memory accesses are not bounds-checked, but the mutations which construct programs are designed never to insert invalid accesses.

The \textit{PRNG state}, the \textit{per-node storage}, and the \textit{graph representation} are all accessed, used, and modified by dedicated instructions. In the case of the per-node storage and the graph, the contents of registers are used as addresses, so we introduce bounds-checking. Invalid write accesses do nothing, while invalid reads return 0.

\paragraph{Graph format.}
The details of the graph storage format require special attention. Since we are interested in being able to add and remove edges quickly ($O(1)$), it is best to represent the graph through its adjacency matrix. Unfortunately, the naive approach of storing the complete matrix has a space requirement of $O(N^2)$, where $N$ is the number of nodes, which becomes prohibitive for large graphs. Our solution is to make use of the sparsity structure of the graphs of interest
and employ a sparse-matrix format. The format we choose is based on the dictionary of keys. We use a set based on a hash table to store the row-column index pairs of non-zero entries in the matrix, giving us amortized $O(1)$ performance for adding and removing edges. In addition, to enforce the symmetry of the matrix (remember that we are only working with undirected graphs), we sort the indices in each index pair so that only the elements above the main diagonal have to be stored.

\paragraph{Instructions.}
With the structure of the memory explained, let us shift our focus to the instructions. As mentioned earlier, an instruction consists of an op-code and operands. Specifically, we represent an instruction as a struct with one op-code, one output operand, and two input operands. Each of these components uses 4 bytes. Depending on the op-code, the operands are interpreted as floating point constants, integer constants, or addresses (which are also integers). The op-code also dictates the type of register towards which an address points.

The complete list of op-codes is presented in Table~\ref{tab:opcodes}. Generally, the suffix indicates the input type, except for \texttt{B\_TO\_I} (conversion from bool to int), and the random sampling functions. Op-codes that contain the ``\texttt{\_IMM\_}'' substring are operations where one of the input operands, usually the second, is a constant (an ``immediate'' operand). One crucial design decision is how to respond to invalid operations, such as division by zeros. Since we expect it is hard to generate code that has no such operations, we decided that they will simply be skipped when encountered during execution.

\begin{table}
    \scriptsize
    \centering
    \begin{tabular}{p{34mm}l}
\toprule
Type & Operations  \\
\midrule
\multirow{2}{3.4cm}{Integer arithmetic}
                                     & PLUS\_I, MINUS\_I, TIMES\_I, DIV\_I, MOD\_I \\
                                     & PLUS\_IMM\_I, MINUS\_IMM\_I, TIMES\_IMM\_I, DIV\_IMM\_I, MOD\_IMM\_I \\
                                     \cmidrule(lr){2-2}
\multirow{3}{3.4cm}{Integer relational operations}
                                     & LT\_I, LTE\_I, EQ\_I, GTE\_I, GT\_I, NEQ\_I \\
                                     & LT\_IMM\_I, LTE\_IMM\_I, EQ\_IMM\_I, GTE\_IMM\_I, GT\_IMM\_I, NEQ\_IMM\_I \\
                                     & NZERO\_I, ZERO\_I \\
                                     \cmidrule(lr){2-2}
\multirow{2}{3.4cm}{Floating-point relational operations}
                                     & LT\_F, LTE\_F, GTE\_F, GT\_F \\
                                     & LT\_IMM\_F, LTE\_IMM\_F, GTE\_IMM\_F, GT\_IMM\_F \\
                                     \cmidrule(lr){2-2}
Boolean operations                   & AND\_B, NAND\_B, OR\_B, NOR\_B, XOR\_B, XNOR\_B, NOT\_B \\
                                     \cmidrule(lr){2-2}
Conversion operations                & B\_TO\_I \\
                                     \cmidrule(lr){2-2}
\multirow{2}{3.4cm}{Randomness}      & RND\_UNIF\_F, RND\_UNIF\_I \\
                                     & RND\_UNIF\_IMM\_I \\
                                     \cmidrule(lr){2-2}
\multirow{2}{3.4cm}{Assignment}      & ASGN\_I, ASGN\_F, ASGN\_B \\
                                     & CONST\_I, CONST\_F \\
                                     \cmidrule(lr){2-2}
Edge operations                      & ADD\_EDGE, REMOVE\_EDGE, FLIP\_EDGE, IS\_EDGE \\
                                     \cmidrule(lr){2-2}
Jumps                                & JUMP\_ABS, JUMP\_REL, JUMPZ, JUMPNZ \\
                                     \cmidrule(lr){2-2}
Call inner loop function             & CALL \\
                                     \cmidrule(lr){2-2}
Per-node storage access             & STORE\_I, LOAD\_I \\
\bottomrule
\end{tabular}
\caption{Op-codes considered in our search.}
\label{tab:opcodes}
\end{table}

\paragraph{Individual execution.} With the details of the instructions and the memory in mind, running an individual to generate graphs is straightforward. We first do some setup, which consists of compiling the code trees, emptying the memory (values for all possible variables are initially set to 0), and loading all input data into memory. Then, for each node, we iterate over and execute the compiled instructions of the first code tree, which corresponds to the \texttt{main} function in our code listings. To ensure the instructions can access the loop index and the program counter, the loop variables of the for-loops over nodes and instructions are references to entries in the memory object.

Executing a \texttt{CALL} instruction represents a special case. For it, we save the program counter value, execute the \texttt{call()} function in a similar fashion to the current one, and then restore the value of the program counter to continue execution.

At the end of the run, the generated graph is inside the memory object and can be converted to a PyTorch sparse tensor for calculating the loss function.

\section{Evolutionary Search Details}
\label{appx:evo}
In this section, we further detail our evolutionary search setup described in the main paper.

\subsection{Stochastic Tournament Selection}
In classical tournament selection, the winner of a tournament is decided as the individual with the best score, even if that individual is only marginally better than the others in the tournament. This behavior can lead to locally optimal traits spreading too fast through the population and trapping the search. The problem is especially evident if we add a regularization factor to the cost function to direct the search toward shorter programs. In this case, it is challenging for the search to build a complex solution since there is strong selection pressure against adding lines of code that do not lead to an immediate improvement of the cost function.

We propose to generalize the tournament by introducing stochastic sampling to mitigate this issue. For this, we pass the individuals' scores through a softmax function, which induces a probability distribution over the individuals. The softmax has a temperature parameter which, similarly to the learning rate used for other optimization methods, we gradually decrease according to an annealing schedule as the search progresses. We then take one sample from this distribution and declare the resulting individual the winner.

Given individuals $(I_1, \ldots, I_T)$ with associated loss values $(l_1, \ldots, l_T)$, the probability of individual $I_i$ of winning the tournament at temperature $\mathcal{T}>0$ is
\begin{equation}
    p(I_i) = \frac{\exp\left(-\frac{l_i}{\mathcal{T}}\right)}{\sum_{j=0}^T\exp\left(-\frac{l_j}{\mathcal{T}}\right)}.
\end{equation}

As $\mathcal{T} \to \infty$, the distribution tends to the uniform distribution, which removes selection pressure and leads to a random walk through the search space. When $\mathcal{T} \to 0$, the sampling procedure reverts to being an argmin over the losses, with the difference that it samples uniformly over the best-performing individuals if more than one achieves the lowest loss.

By following an annealing schedule that decreases from a high value of $\mathcal{T}$ to 0, we can explore larger parts of the loss landscape in the beginning and switch to an exploiting and fine-tuning behavior towards the end. 

Since it uses the score differences between individuals and not just the rank order, the softmax also lets us combine multiple cost functions in useful ways. In particular, we can add a regularization loss that penalizes long programs to the main loss. By choosing a small scaling factor for the regularization loss, we enable the search to explore programs of all sizes, but still, have a preference for shorter programs with equivalent functionality when they are discovered.

\subsection{Mutations}
The choice of mutation operations dictates how easy it is to transition between candidate solutions and therefore is critical for the performance of the search. We base some of our mutations on those used by \citet{realAutoMLZeroEvolvingMachine2020}, but adapt them to our search space and implement several others.

The list of mutations is as follows:
\begin{enumerate}
    \item Insertion Mutation: insert a random line of code at a random location in the program.
    \item Knockout Mutation: remove a random line of code from the program.
    \item Operation Change Mutation: select a random line of code and replace its operation with a new random one with the same inputs and outputs.
    \item Parameter Change Mutation: select a random line and change one of the input or output arguments with a new random one.
    \item Randomization Mutation: randomize the contents of the program while keeping the size of each function, \texttt{main} and \texttt{call}, the same.
    \item No-Op Mutation: leave the program as it is.
\end{enumerate}

We sample the mutation types at different rates. We chose knockout mutation at triple the baseline rate, while the parameter change mutation happens at double the baseline rate. When a mutation type is sampled, we sample the particular change uniformly at random, from the available ones for that mutation type (e.g. one line is removed uniformly at random). There are two exceptions to this: (1) when we are inserting a line, we choose to insert an if-statement with a probability of 0.2, while otherwise, we insert a random command (Table~\ref{tab:opcodes}); (2) when we are updating a parameter value, we use Brownian motion to ensure that parameter change is gradual. For integer variables, we use the discrete Gaussian distribution with a standard deviation of $1.0$ for the Brownian motion, while for floating point variables we use the Gaussian distribution with a standard deviation of $0.1$.

\subsection{Fitness}
\label{appx:fitness}
As with all optimization procedures, the choice of the loss function is one of the most important design choices and is decisive for the algorithm's performance.

To evaluate an individual's performance, we sample batches of $B$ examples from the dataset, iterate over them and run the individual once for each, using the graph size and the auxiliary data as input. The functions that the individual can call for generating random numbers make use of a random number generator state that is not reset between runs. This implies that the resulting graphs can be considered a random variable with an associated probability distribution. By comparing this probability distribution to the distribution of the graphs in the batch, we can calculate a score for the individual.

As mentioned before, we use MMD to compute the difference between the distributions:
\begin{equation}
    \mathrm{MMD}^2_b\left[k,X,Y\right] = \frac{1}{m^2}\sum_{i,j=1}^{m}k(x_i, x_j) - \frac{2}{mn}\sum_{i,j=1}^{m,n}k(x_i,y_j)+\frac{1}{n^2}\sum_{i,j=1}^{n}k(y_i,y_j),
\end{equation}
where $X$ and $Y$ are samples from two distributions, and $k: \mathcal{X} \times \mathcal{X} \to \mathbb{R}$ is a kernel function \citep{grettonKernelTwoSampleTest2012}.

We follow the usual practice of setting $\mathcal{X} = \mathbb{R}^d$ and applying a feature extraction function $\phi: \mathcal{G} \to \mathbb{R}^d$ to the reference and predicted graphs and using the MMD on the resulting embeddings.

The main feature extraction step that we use is a randomly initialized, untrained GIN neural network \citep{xuHowPowerfulAre2019}. We set the number of message passing rounds to 3, the number of layers in each MLP to 2, and the embedding dimension to 35, as suggested by~\citet{thompsonEvaluationMetricsGraph2022}. As node input features, we use the one-hot encoding of the node degrees after clamping them to the range $[0, 19]$. We deviate from the original GIN model in three ways.

First, we remove the batch normalization layers. Untrained batch norm layers default to a mean of 0 and a variance of 1, so they act as no-ops in our model.

Second, we do not use biases in the MLP layers. Since they were untrained and our input (a one-hot encoding) can never be zero, we considered them redundant. An advantage of doing this is that, when padding the graphs and embedding vectors with zeros to match a fixed graph size, zero elements stay unmodified throughout the GIN's processing, which leads to easier pooling over the graph.

Third, we add a custom normalization step after each messag-passing round. Concretely, for each node, we divide the results of message aggregation by the square root of the number of neighbors. The goal is to reduce the output variance and compensate for the fact that the batch norm layers cannot be used.

With the developed framework, we also have the option to use classical feature extractors, such as histograms of node degrees, histograms of the eigenvalues of the graph Laplacian, and histograms of node clustering coefficients. 
In principle, if one would assume that the ordering of the graphs is known, a powerful alternative feature extraction would be to directly compute the Hamming distance between the generated adjacency matrix and the true adjacency matrix. However, in our experiments, we assume that the ordering is unknown, and effectively leave it to the search procedure to find a node ordering with which it is easy to build a graph construction algorithm.

As kernel functions for the MMD, we implement the Gaussian kernel, also known as the radial basis function kernel:
\begin{align}
    \label{eq:}
    k_{Gauss}(x,y) &= \exp\left(-\frac{\|x-y\|_2^2}{2\sigma^2}\right) \\
\end{align}

\subsection{Hyperparameters}
\label{appx:hyperparams}

For each dataset, we train models that have access to all operations, with a population of $1000$ individuals, tournaments of size $4$, and keep an elite set of $10$ individuals when performing the generational replacement. We anneal the temperature exponentially from a value of $10$ towards $0$ with a factor of $0.9998$ per generation. 
The cost function is the MMD with a GIN feature extractor and using the Gaussian kernel with $\sigma=1$, onto which we add a regularization cost of $10^{-8}$ for each node in an individual's code trees. The code length is capped to a maximum of $50$ nodes in the tree. All experiments are run on multi-core CPU machines without GPU acceleration. The search experiments are run for $89$ hours or $1$ million generations, whichever comes first.

\section{Examples of Discovered Algorithms}\label{appx:discovered_algos}
Here in Figure~\ref{fig:algs} we show the algorithms our method discovered for the graph classes we have considered. We can see that the discovered algorithms are mostly interpretable, except the algorithm for grids (Figure~\ref{fig:alg_grids}). Which likely grows in complexity because it is quite hard to determine a possible width of a grid from the desired number of nodes without having a function for number factorization. The function discovered for lobster graph construction produces graphs very close to lobster graphs, as the nodes with low IDs are prioritized for attachment, forming a sort of a backbone. Hover the strict limit of having at most two children outside of the backbone was not discovered. This can potentially be alleviated by performing the search with more individuals and running it for longer. However, we did not test this due to the time limit.
The search procedure only ended up using one loop for these datasets, as we apply some pressure on the program length and the solutions with two loops did not prove to be better. This can be adjusted, either by decreasing the pressure on program length or forcing it to execute the inner loop.

\begin{figure}[hpt]
    \lstset{style=modelstyle}
    \begin{minipage}{\linewidth}
    \centering
    \begin{subfigure}[]{0.35\textwidth}
        \begin{center}
            \begin{lstlisting}[language=Python, basicstyle=\ttfamily\tiny,gobble=16]
                def main():
                  for i in range(N):
                    flip_edge(int03, i)
                    int03 = uniform(i)
            \end{lstlisting}
        \end{center}
        \caption{Protein}
        \label{fig:alg_proteins}
    \end{subfigure}
    \hspace{2em}
    \begin{subfigure}[]{0.35\textwidth}
        \begin{center}
            \begin{lstlisting}[language=Python, basicstyle=\ttfamily\tiny,gobble=16]
                def main():
                  for i in range(N):
                    int05 = uniform(i)
                    int08 = int05 // 2
                    add_edge(i, int08)
            \end{lstlisting}
        \end{center}
        \caption{Lobster}
        \label{fig:alg_lobsters}
    \end{subfigure}
    \\
    \begin{subfigure}[]{0.35\textwidth}
        \begin{center}
            \begin{lstlisting}[language=Python, basicstyle=\ttfamily\tiny,gobble=16]
                def main():
                  for i in range(N):
                    remove_edge(int06, int10)
                    add_edge(i, int06)
                    int07 = i % W
                    int09 = i - W
                    add_edge(int09, i)
                    int10 = int07 + int06
                    int06 = i + int05
            \end{lstlisting}
        \end{center}
        \caption{Grid with provided width}
        \label{fig:alg_grids_w_input}
    \end{subfigure}
    \hspace{2em}
    \begin{subfigure}[]{0.35\textwidth}
        \begin{center}
        \vspace{14pt}
            \begin{lstlisting}[language=Python, basicstyle=\ttfamily\tiny,gobble=16]
                def main():
                  for i in range(N):
                    bool05 = int07 >= 0
                    int04 = i - int08
                    if bool04:
                      bool05 = int07 >= int06
                      int03 = int08 + 3
                      int07 -= 2
                    add_edge(i, int03)
                    bool02 = bool05 or bool02
                    if not bool02:
                      int06 = int05 + 1
                      flip_edge(int08, int04)
                    int08 = int05 - int04
                    int03 = i
                    bool00 = int06 < 1
                    flip_edge(int06, i)
                    remove_edge(int08, int06)
                    int08 = int07 + 4
                    int06 /= int05
                    if bool00:
                      int07 = int05 + int03
                      int08 = j
                      int04 = i // 5
                      int06 = j // 3
                      bool05 = is_edge(N, i)
                      bool01 = int04 > int05
                      int05 = int04 * int07
                    else:
                      int07 = int05 + N
                      if bool01:
                        int08 += 5
                        int04 = 5
                        remove_edge(int08, int07)
                        int03 = 5
                      bool01 = int06 >= 4
                    bool02 = int06 < int04
                    int05 += 3
                    if bool01:
                      int05 = int04 // 3
                      add_edge(int05, int04)
                    else:
                      bool04 = int05 == int06
                      int05 = int03 % int06
                    int06 = int03 - 5
                    bool00 = int04 >= int05
                    if not bool05:
                      bool01 = is_edge(int06, int08)
                      if bool01:
                        int03 = int03 % i
                      bool04 = bool00 and bool05
                    bool02 = bool00 != bool02
                    int05 = int04 + int03
            \end{lstlisting}
        \end{center}
        \caption{Grid}
        \label{fig:alg_grids}
    \end{subfigure}
    \end{minipage}
    \lstset{style=mystyle}
    \caption{Generated algorithms for each experiment. Algorithms for \nameref{fig:alg_lobsters} and \nameref{fig:alg_grids} were manually cleaned, by removing redundant lines.}
    \label{fig:algs}
\end{figure}

\end{document}

%% file: math_commands.tex
\usepackage{amsmath,amsfonts,bm}

\def\eqref#1{equation~\ref{#1}}

\def\1{\bm{1}}

\DeclareMathAlphabet{\mathsfit}{\encodingdefault}{\sfdefault}{m}{sl}
\SetMathAlphabet{\mathsfit}{bold}{\encodingdefault}{\sfdefault}{bx}{n}



%% file: figures/true_Grids.tex
%

\begin{tikzpicture}[anchor=mid,>=latex',]
\pgfsetlinewidth{1bp}
\node (0) at (98.064bp,18.0bp) [draw=graphnodecolor,fill=graphnodecolor,circle] {};
  \node (1) at (178.77bp,34.179bp) [draw=graphnodecolor,fill=graphnodecolor,circle] {};
  \node (9) at (79.006bp,100.42bp) [draw=graphnodecolor,fill=graphnodecolor,circle] {};
  \node (2) at (260.83bp,51.421bp) [draw=graphnodecolor,fill=graphnodecolor,circle] {};
  \node (10) at (160.1bp,118.74bp) [draw=graphnodecolor,fill=graphnodecolor,circle] {};
  \node (3) at (343.06bp,69.72bp) [draw=graphnodecolor,fill=graphnodecolor,circle] {};
  \node (11) at (241.97bp,137.18bp) [draw=graphnodecolor,fill=graphnodecolor,circle] {};
  \node (4) at (425.19bp,88.733bp) [draw=graphnodecolor,fill=graphnodecolor,circle] {};
  \node (12) at (323.55bp,156.05bp) [draw=graphnodecolor,fill=graphnodecolor,circle] {};
  \node (5) at (507.18bp,108.32bp) [draw=graphnodecolor,fill=graphnodecolor,circle] {};
  \node (13) at (404.91bp,175.11bp) [draw=graphnodecolor,fill=graphnodecolor,circle] {};
  \node (6) at (588.93bp,128.52bp) [draw=graphnodecolor,fill=graphnodecolor,circle] {};
  \node (14) at (486.23bp,194.33bp) [draw=graphnodecolor,fill=graphnodecolor,circle] {};
  \node (7) at (670.08bp,149.53bp) [draw=graphnodecolor,fill=graphnodecolor,circle] {};
  \node (15) at (567.68bp,213.74bp) [draw=graphnodecolor,fill=graphnodecolor,circle] {};
  \node (8) at (749.56bp,170.82bp) [draw=graphnodecolor,fill=graphnodecolor,circle] {};
  \node (16) at (649.21bp,233.58bp) [draw=graphnodecolor,fill=graphnodecolor,circle] {};
  \node (17) at (730.0bp,253.13bp) [draw=graphnodecolor,fill=graphnodecolor,circle] {};
  \node (18) at (58.756bp,184.57bp) [draw=graphnodecolor,fill=graphnodecolor,circle] {};
  \node (19) at (139.69bp,204.57bp) [draw=graphnodecolor,fill=graphnodecolor,circle] {};
  \node (20) at (221.39bp,224.67bp) [draw=graphnodecolor,fill=graphnodecolor,circle] {};
  \node (21) at (302.82bp,244.48bp) [draw=graphnodecolor,fill=graphnodecolor,circle] {};
  \node (22) at (384.08bp,263.84bp) [draw=graphnodecolor,fill=graphnodecolor,circle] {};
  \node (23) at (465.46bp,282.75bp) [draw=graphnodecolor,fill=graphnodecolor,circle] {};
  \node (24) at (547.18bp,301.25bp) [draw=graphnodecolor,fill=graphnodecolor,circle] {};
  \node (25) at (629.31bp,319.53bp) [draw=graphnodecolor,fill=graphnodecolor,circle] {};
  \node (26) at (710.72bp,337.52bp) [draw=graphnodecolor,fill=graphnodecolor,circle] {};
  \node (27) at (38.664bp,268.7bp) [draw=graphnodecolor,fill=graphnodecolor,circle] {};
  \node (28) at (119.12bp,290.33bp) [draw=graphnodecolor,fill=graphnodecolor,circle] {};
  \node (29) at (200.41bp,312.0bp) [draw=graphnodecolor,fill=graphnodecolor,circle] {};
  \node (30) at (281.79bp,332.75bp) [draw=graphnodecolor,fill=graphnodecolor,circle] {};
  \node (31) at (363.28bp,352.46bp) [draw=graphnodecolor,fill=graphnodecolor,circle] {};
  \node (32) at (445.01bp,371.15bp) [draw=graphnodecolor,fill=graphnodecolor,circle] {};
  \node (33) at (527.12bp,388.8bp) [draw=graphnodecolor,fill=graphnodecolor,circle] {};
  \node (34) at (609.58bp,405.49bp) [draw=graphnodecolor,fill=graphnodecolor,circle] {};
  \node (35) at (691.31bp,421.8bp) [draw=graphnodecolor,fill=graphnodecolor,circle] {};
  \node (36) at (18.0bp,350.59bp) [draw=graphnodecolor,fill=graphnodecolor,circle] {};
  \node (37) at (97.272bp,373.97bp) [draw=graphnodecolor,fill=graphnodecolor,circle] {};
  \node (38) at (178.45bp,396.78bp) [draw=graphnodecolor,fill=graphnodecolor,circle] {};
  \node (39) at (260.51bp,418.32bp) [draw=graphnodecolor,fill=graphnodecolor,circle] {};
  \node (40) at (343.08bp,438.47bp) [draw=graphnodecolor,fill=graphnodecolor,circle] {};
  \node (41) at (425.99bp,457.23bp) [draw=graphnodecolor,fill=graphnodecolor,circle] {};
  \node (42) at (509.09bp,474.48bp) [draw=graphnodecolor,fill=graphnodecolor,circle] {};
  \node (43) at (591.99bp,490.11bp) [draw=graphnodecolor,fill=graphnodecolor,circle] {};
  \node (44) at (673.43bp,504.33bp) [draw=graphnodecolor,fill=graphnodecolor,circle] {};
  \draw [] (0) ..controls (129.35bp,24.272bp) and (147.55bp,27.92bp)  .. (1);
  \draw [] (0) ..controls (90.71bp,49.803bp) and (86.267bp,69.016bp)  .. (9);
  \draw [] (1) ..controls (210.34bp,40.813bp) and (229.3bp,44.797bp)  .. (2);
  \draw [] (1) ..controls (171.62bp,66.558bp) and (167.16bp,86.758bp)  .. (10);
  \draw [] (2) ..controls (292.46bp,58.461bp) and (311.46bp,62.69bp)  .. (3);
  \draw [] (2) ..controls (253.68bp,83.91bp) and (249.11bp,104.72bp)  .. (11);
  \draw [] (3) ..controls (374.65bp,77.035bp) and (393.63bp,81.428bp)  .. (4);
  \draw [] (3) ..controls (335.64bp,102.52bp) and (330.87bp,123.65bp)  .. (12);
  \draw [] (4) ..controls (456.73bp,96.267bp) and (475.68bp,100.79bp)  .. (5);
  \draw [] (4) ..controls (417.48bp,121.55bp) and (412.52bp,142.7bp)  .. (13);
  \draw [] (5) ..controls (538.63bp,116.09bp) and (557.52bp,120.76bp)  .. (6);
  \draw [] (5) ..controls (499.22bp,141.0bp) and (494.09bp,162.05bp)  .. (14);
  \draw [] (6) ..controls (620.15bp,136.6bp) and (638.9bp,141.46bp)  .. (7);
  \draw [] (6) ..controls (580.88bp,160.8bp) and (575.72bp,181.49bp)  .. (15);
  \draw [] (7) ..controls (700.89bp,157.78bp) and (718.81bp,162.58bp)  .. (8);
  \draw [] (7) ..controls (662.09bp,181.71bp) and (657.1bp,201.79bp)  .. (16);
  \draw [] (8) ..controls (742.01bp,202.58bp) and (737.45bp,221.77bp)  .. (17);
  \draw [] (9) ..controls (110.2bp,107.47bp) and (128.94bp,111.7bp)  .. (10);
  \draw [] (9) ..controls (71.252bp,132.64bp) and (66.414bp,152.75bp)  .. (18);
  \draw [] (10) ..controls (191.6bp,125.83bp) and (210.52bp,130.1bp)  .. (11);
  \draw [] (10) ..controls (152.37bp,151.25bp) and (147.41bp,172.08bp)  .. (19);
  \draw [] (11) ..controls (273.36bp,144.44bp) and (292.21bp,148.8bp)  .. (12);
  \draw [] (11) ..controls (234.24bp,170.07bp) and (229.12bp,191.82bp)  .. (20);
  \draw [] (12) ..controls (354.85bp,163.38bp) and (373.65bp,167.79bp)  .. (13);
  \draw [] (12) ..controls (315.73bp,189.39bp) and (310.53bp,211.56bp)  .. (21);
  \draw [] (13) ..controls (436.2bp,182.5bp) and (454.99bp,186.94bp)  .. (14);
  \draw [] (13) ..controls (397.17bp,208.09bp) and (391.91bp,230.51bp)  .. (22);
  \draw [] (14) ..controls (517.56bp,201.8bp) and (536.38bp,206.28bp)  .. (15);
  \draw [] (14) ..controls (478.4bp,227.66bp) and (473.19bp,249.84bp)  .. (23);
  \draw [] (15) ..controls (599.05bp,221.37bp) and (617.89bp,225.96bp)  .. (16);
  \draw [] (15) ..controls (559.97bp,246.63bp) and (554.88bp,268.39bp)  .. (24);
  \draw [] (16) ..controls (680.3bp,241.1bp) and (698.96bp,245.62bp)  .. (17);
  \draw [] (16) ..controls (641.67bp,266.14bp) and (636.84bp,287.0bp)  .. (25);
  \draw [] (17) ..controls (722.62bp,285.44bp) and (718.01bp,305.6bp)  .. (26);
  \draw [] (18) ..controls (89.893bp,192.27bp) and (108.6bp,196.89bp)  .. (19);
  \draw [] (18) ..controls (51.062bp,216.79bp) and (46.263bp,236.88bp)  .. (27);
  \draw [] (19) ..controls (171.12bp,212.31bp) and (190.0bp,216.95bp)  .. (20);
  \draw [] (19) ..controls (131.9bp,237.06bp) and (126.91bp,257.87bp)  .. (28);
  \draw [] (20) ..controls (252.72bp,232.29bp) and (271.53bp,236.87bp)  .. (21);
  \draw [] (20) ..controls (213.51bp,257.5bp) and (208.29bp,279.21bp)  .. (29);
  \draw [] (21) ..controls (334.08bp,251.93bp) and (352.86bp,256.4bp)  .. (22);
  \draw [] (21) ..controls (294.89bp,277.76bp) and (289.61bp,299.9bp)  .. (30);
  \draw [] (22) ..controls (415.39bp,271.11bp) and (434.19bp,275.48bp)  .. (23);
  \draw [] (22) ..controls (376.24bp,297.25bp) and (371.02bp,319.48bp)  .. (31);
  \draw [] (23) ..controls (496.9bp,289.87bp) and (515.78bp,294.14bp)  .. (24);
  \draw [] (23) ..controls (457.75bp,316.07bp) and (452.62bp,338.25bp)  .. (32);
  \draw [] (24) ..controls (578.78bp,308.28bp) and (597.76bp,312.51bp)  .. (25);
  \draw [] (24) ..controls (539.64bp,334.16bp) and (534.66bp,355.93bp)  .. (33);
  \draw [] (25) ..controls (660.63bp,326.45bp) and (679.44bp,330.61bp)  .. (26);
  \draw [] (25) ..controls (621.84bp,352.09bp) and (617.05bp,372.96bp)  .. (34);
  \draw [] (26) ..controls (703.29bp,369.79bp) and (698.65bp,389.92bp)  .. (35);
  \draw [] (27) ..controls (69.618bp,277.02bp) and (88.211bp,282.02bp)  .. (28);
  \draw [] (27) ..controls (30.714bp,300.21bp) and (25.939bp,319.13bp)  .. (36);
  \draw [] (28) ..controls (150.4bp,298.67bp) and (169.18bp,303.68bp)  .. (29);
  \draw [] (28) ..controls (110.76bp,322.36bp) and (105.54bp,342.34bp)  .. (37);
  \draw [] (29) ..controls (231.72bp,319.99bp) and (250.52bp,324.78bp)  .. (30);
  \draw [] (29) ..controls (192.09bp,344.12bp) and (186.76bp,364.7bp)  .. (38);
  \draw [] (30) ..controls (313.14bp,340.33bp) and (331.97bp,344.89bp)  .. (31);
  \draw [] (30) ..controls (273.73bp,365.17bp) and (268.56bp,385.93bp)  .. (39);
  \draw [] (31) ..controls (394.72bp,359.65bp) and (413.61bp,363.97bp)  .. (32);
  \draw [] (31) ..controls (355.63bp,385.04bp) and (350.73bp,405.92bp)  .. (40);
  \draw [] (32) ..controls (476.6bp,377.94bp) and (495.58bp,382.02bp)  .. (33);
  \draw [] (32) ..controls (437.81bp,403.76bp) and (433.19bp,424.65bp)  .. (41);
  \draw [] (33) ..controls (558.85bp,395.22bp) and (577.9bp,399.08bp)  .. (34);
  \draw [] (33) ..controls (520.29bp,421.26bp) and (515.92bp,442.05bp)  .. (42);
  \draw [] (34) ..controls (641.02bp,411.76bp) and (659.91bp,415.53bp)  .. (35);
  \draw [] (34) ..controls (602.85bp,437.89bp) and (598.64bp,458.11bp)  .. (43);
  \draw [] (35) ..controls (684.43bp,453.55bp) and (680.3bp,472.62bp)  .. (44);
  \draw [] (36) ..controls (48.821bp,359.68bp) and (66.852bp,365.0bp)  .. (37);
  \draw [] (37) ..controls (128.5bp,382.74bp) and (147.26bp,388.02bp)  .. (38);
  \draw [] (38) ..controls (210.11bp,405.09bp) and (229.24bp,410.11bp)  .. (39);
  \draw [] (39) ..controls (291.94bp,425.99bp) and (311.33bp,430.72bp)  .. (40);
  \draw [] (40) ..controls (374.74bp,445.63bp) and (394.37bp,450.08bp)  .. (41);
  \draw [] (41) ..controls (458.06bp,463.88bp) and (477.43bp,467.91bp)  .. (42);
  \draw [] (42) ..controls (540.98bp,480.49bp) and (560.14bp,484.1bp)  .. (43);
  \draw [] (43) ..controls (623.65bp,495.64bp) and (642.18bp,498.87bp)  .. (44);
\end{tikzpicture}

%

%% file: figures/true_Grids_with_info.tex
%

\begin{tikzpicture}[anchor=mid,>=latex',]
\pgfsetlinewidth{1bp}
\node (0) at (98.064bp,18.0bp) [draw=graphnodecolor,fill=graphnodecolor,circle] {};
  \node (1) at (178.77bp,34.179bp) [draw=graphnodecolor,fill=graphnodecolor,circle] {};
  \node (9) at (79.006bp,100.42bp) [draw=graphnodecolor,fill=graphnodecolor,circle] {};
  \node (2) at (260.83bp,51.421bp) [draw=graphnodecolor,fill=graphnodecolor,circle] {};
  \node (10) at (160.1bp,118.74bp) [draw=graphnodecolor,fill=graphnodecolor,circle] {};
  \node (3) at (343.06bp,69.72bp) [draw=graphnodecolor,fill=graphnodecolor,circle] {};
  \node (11) at (241.97bp,137.18bp) [draw=graphnodecolor,fill=graphnodecolor,circle] {};
  \node (4) at (425.19bp,88.733bp) [draw=graphnodecolor,fill=graphnodecolor,circle] {};
  \node (12) at (323.55bp,156.05bp) [draw=graphnodecolor,fill=graphnodecolor,circle] {};
  \node (5) at (507.18bp,108.32bp) [draw=graphnodecolor,fill=graphnodecolor,circle] {};
  \node (13) at (404.91bp,175.11bp) [draw=graphnodecolor,fill=graphnodecolor,circle] {};
  \node (6) at (588.93bp,128.52bp) [draw=graphnodecolor,fill=graphnodecolor,circle] {};
  \node (14) at (486.23bp,194.33bp) [draw=graphnodecolor,fill=graphnodecolor,circle] {};
  \node (7) at (670.08bp,149.53bp) [draw=graphnodecolor,fill=graphnodecolor,circle] {};
  \node (15) at (567.68bp,213.74bp) [draw=graphnodecolor,fill=graphnodecolor,circle] {};
  \node (8) at (749.56bp,170.82bp) [draw=graphnodecolor,fill=graphnodecolor,circle] {};
  \node (16) at (649.21bp,233.58bp) [draw=graphnodecolor,fill=graphnodecolor,circle] {};
  \node (17) at (730.0bp,253.13bp) [draw=graphnodecolor,fill=graphnodecolor,circle] {};
  \node (18) at (58.756bp,184.57bp) [draw=graphnodecolor,fill=graphnodecolor,circle] {};
  \node (19) at (139.69bp,204.57bp) [draw=graphnodecolor,fill=graphnodecolor,circle] {};
  \node (20) at (221.39bp,224.67bp) [draw=graphnodecolor,fill=graphnodecolor,circle] {};
  \node (21) at (302.82bp,244.48bp) [draw=graphnodecolor,fill=graphnodecolor,circle] {};
  \node (22) at (384.08bp,263.84bp) [draw=graphnodecolor,fill=graphnodecolor,circle] {};
  \node (23) at (465.46bp,282.75bp) [draw=graphnodecolor,fill=graphnodecolor,circle] {};
  \node (24) at (547.18bp,301.25bp) [draw=graphnodecolor,fill=graphnodecolor,circle] {};
  \node (25) at (629.31bp,319.53bp) [draw=graphnodecolor,fill=graphnodecolor,circle] {};
  \node (26) at (710.72bp,337.52bp) [draw=graphnodecolor,fill=graphnodecolor,circle] {};
  \node (27) at (38.664bp,268.7bp) [draw=graphnodecolor,fill=graphnodecolor,circle] {};
  \node (28) at (119.12bp,290.33bp) [draw=graphnodecolor,fill=graphnodecolor,circle] {};
  \node (29) at (200.41bp,312.0bp) [draw=graphnodecolor,fill=graphnodecolor,circle] {};
  \node (30) at (281.79bp,332.75bp) [draw=graphnodecolor,fill=graphnodecolor,circle] {};
  \node (31) at (363.28bp,352.46bp) [draw=graphnodecolor,fill=graphnodecolor,circle] {};
  \node (32) at (445.01bp,371.15bp) [draw=graphnodecolor,fill=graphnodecolor,circle] {};
  \node (33) at (527.12bp,388.8bp) [draw=graphnodecolor,fill=graphnodecolor,circle] {};
  \node (34) at (609.58bp,405.49bp) [draw=graphnodecolor,fill=graphnodecolor,circle] {};
  \node (35) at (691.31bp,421.8bp) [draw=graphnodecolor,fill=graphnodecolor,circle] {};
  \node (36) at (18.0bp,350.59bp) [draw=graphnodecolor,fill=graphnodecolor,circle] {};
  \node (37) at (97.272bp,373.97bp) [draw=graphnodecolor,fill=graphnodecolor,circle] {};
  \node (38) at (178.45bp,396.78bp) [draw=graphnodecolor,fill=graphnodecolor,circle] {};
  \node (39) at (260.51bp,418.32bp) [draw=graphnodecolor,fill=graphnodecolor,circle] {};
  \node (40) at (343.08bp,438.47bp) [draw=graphnodecolor,fill=graphnodecolor,circle] {};
  \node (41) at (425.99bp,457.23bp) [draw=graphnodecolor,fill=graphnodecolor,circle] {};
  \node (42) at (509.09bp,474.48bp) [draw=graphnodecolor,fill=graphnodecolor,circle] {};
  \node (43) at (591.99bp,490.11bp) [draw=graphnodecolor,fill=graphnodecolor,circle] {};
  \node (44) at (673.43bp,504.33bp) [draw=graphnodecolor,fill=graphnodecolor,circle] {};
  \draw [] (0) ..controls (129.35bp,24.272bp) and (147.55bp,27.92bp)  .. (1);
  \draw [] (0) ..controls (90.71bp,49.803bp) and (86.267bp,69.016bp)  .. (9);
  \draw [] (1) ..controls (210.34bp,40.813bp) and (229.3bp,44.797bp)  .. (2);
  \draw [] (1) ..controls (171.62bp,66.558bp) and (167.16bp,86.758bp)  .. (10);
  \draw [] (2) ..controls (292.46bp,58.461bp) and (311.46bp,62.69bp)  .. (3);
  \draw [] (2) ..controls (253.68bp,83.91bp) and (249.11bp,104.72bp)  .. (11);
  \draw [] (3) ..controls (374.65bp,77.035bp) and (393.63bp,81.428bp)  .. (4);
  \draw [] (3) ..controls (335.64bp,102.52bp) and (330.87bp,123.65bp)  .. (12);
  \draw [] (4) ..controls (456.73bp,96.267bp) and (475.68bp,100.79bp)  .. (5);
  \draw [] (4) ..controls (417.48bp,121.55bp) and (412.52bp,142.7bp)  .. (13);
  \draw [] (5) ..controls (538.63bp,116.09bp) and (557.52bp,120.76bp)  .. (6);
  \draw [] (5) ..controls (499.22bp,141.0bp) and (494.09bp,162.05bp)  .. (14);
  \draw [] (6) ..controls (620.15bp,136.6bp) and (638.9bp,141.46bp)  .. (7);
  \draw [] (6) ..controls (580.88bp,160.8bp) and (575.72bp,181.49bp)  .. (15);
  \draw [] (7) ..controls (700.89bp,157.78bp) and (718.81bp,162.58bp)  .. (8);
  \draw [] (7) ..controls (662.09bp,181.71bp) and (657.1bp,201.79bp)  .. (16);
  \draw [] (8) ..controls (742.01bp,202.58bp) and (737.45bp,221.77bp)  .. (17);
  \draw [] (9) ..controls (110.2bp,107.47bp) and (128.94bp,111.7bp)  .. (10);
  \draw [] (9) ..controls (71.252bp,132.64bp) and (66.414bp,152.75bp)  .. (18);
  \draw [] (10) ..controls (191.6bp,125.83bp) and (210.52bp,130.1bp)  .. (11);
  \draw [] (10) ..controls (152.37bp,151.25bp) and (147.41bp,172.08bp)  .. (19);
  \draw [] (11) ..controls (273.36bp,144.44bp) and (292.21bp,148.8bp)  .. (12);
  \draw [] (11) ..controls (234.24bp,170.07bp) and (229.12bp,191.82bp)  .. (20);
  \draw [] (12) ..controls (354.85bp,163.38bp) and (373.65bp,167.79bp)  .. (13);
  \draw [] (12) ..controls (315.73bp,189.39bp) and (310.53bp,211.56bp)  .. (21);
  \draw [] (13) ..controls (436.2bp,182.5bp) and (454.99bp,186.94bp)  .. (14);
  \draw [] (13) ..controls (397.17bp,208.09bp) and (391.91bp,230.51bp)  .. (22);
  \draw [] (14) ..controls (517.56bp,201.8bp) and (536.38bp,206.28bp)  .. (15);
  \draw [] (14) ..controls (478.4bp,227.66bp) and (473.19bp,249.84bp)  .. (23);
  \draw [] (15) ..controls (599.05bp,221.37bp) and (617.89bp,225.96bp)  .. (16);
  \draw [] (15) ..controls (559.97bp,246.63bp) and (554.88bp,268.39bp)  .. (24);
  \draw [] (16) ..controls (680.3bp,241.1bp) and (698.96bp,245.62bp)  .. (17);
  \draw [] (16) ..controls (641.67bp,266.14bp) and (636.84bp,287.0bp)  .. (25);
  \draw [] (17) ..controls (722.62bp,285.44bp) and (718.01bp,305.6bp)  .. (26);
  \draw [] (18) ..controls (89.893bp,192.27bp) and (108.6bp,196.89bp)  .. (19);
  \draw [] (18) ..controls (51.062bp,216.79bp) and (46.263bp,236.88bp)  .. (27);
  \draw [] (19) ..controls (171.12bp,212.31bp) and (190.0bp,216.95bp)  .. (20);
  \draw [] (19) ..controls (131.9bp,237.06bp) and (126.91bp,257.87bp)  .. (28);
  \draw [] (20) ..controls (252.72bp,232.29bp) and (271.53bp,236.87bp)  .. (21);
  \draw [] (20) ..controls (213.51bp,257.5bp) and (208.29bp,279.21bp)  .. (29);
  \draw [] (21) ..controls (334.08bp,251.93bp) and (352.86bp,256.4bp)  .. (22);
  \draw [] (21) ..controls (294.89bp,277.76bp) and (289.61bp,299.9bp)  .. (30);
  \draw [] (22) ..controls (415.39bp,271.11bp) and (434.19bp,275.48bp)  .. (23);
  \draw [] (22) ..controls (376.24bp,297.25bp) and (371.02bp,319.48bp)  .. (31);
  \draw [] (23) ..controls (496.9bp,289.87bp) and (515.78bp,294.14bp)  .. (24);
  \draw [] (23) ..controls (457.75bp,316.07bp) and (452.62bp,338.25bp)  .. (32);
  \draw [] (24) ..controls (578.78bp,308.28bp) and (597.76bp,312.51bp)  .. (25);
  \draw [] (24) ..controls (539.64bp,334.16bp) and (534.66bp,355.93bp)  .. (33);
  \draw [] (25) ..controls (660.63bp,326.45bp) and (679.44bp,330.61bp)  .. (26);
  \draw [] (25) ..controls (621.84bp,352.09bp) and (617.05bp,372.96bp)  .. (34);
  \draw [] (26) ..controls (703.29bp,369.79bp) and (698.65bp,389.92bp)  .. (35);
  \draw [] (27) ..controls (69.618bp,277.02bp) and (88.211bp,282.02bp)  .. (28);
  \draw [] (27) ..controls (30.714bp,300.21bp) and (25.939bp,319.13bp)  .. (36);
  \draw [] (28) ..controls (150.4bp,298.67bp) and (169.18bp,303.68bp)  .. (29);
  \draw [] (28) ..controls (110.76bp,322.36bp) and (105.54bp,342.34bp)  .. (37);
  \draw [] (29) ..controls (231.72bp,319.99bp) and (250.52bp,324.78bp)  .. (30);
  \draw [] (29) ..controls (192.09bp,344.12bp) and (186.76bp,364.7bp)  .. (38);
  \draw [] (30) ..controls (313.14bp,340.33bp) and (331.97bp,344.89bp)  .. (31);
  \draw [] (30) ..controls (273.73bp,365.17bp) and (268.56bp,385.93bp)  .. (39);
  \draw [] (31) ..controls (394.72bp,359.65bp) and (413.61bp,363.97bp)  .. (32);
  \draw [] (31) ..controls (355.63bp,385.04bp) and (350.73bp,405.92bp)  .. (40);
  \draw [] (32) ..controls (476.6bp,377.94bp) and (495.58bp,382.02bp)  .. (33);
  \draw [] (32) ..controls (437.81bp,403.76bp) and (433.19bp,424.65bp)  .. (41);
  \draw [] (33) ..controls (558.85bp,395.22bp) and (577.9bp,399.08bp)  .. (34);
  \draw [] (33) ..controls (520.29bp,421.26bp) and (515.92bp,442.05bp)  .. (42);
  \draw [] (34) ..controls (641.02bp,411.76bp) and (659.91bp,415.53bp)  .. (35);
  \draw [] (34) ..controls (602.85bp,437.89bp) and (598.64bp,458.11bp)  .. (43);
  \draw [] (35) ..controls (684.43bp,453.55bp) and (680.3bp,472.62bp)  .. (44);
  \draw [] (36) ..controls (48.821bp,359.68bp) and (66.852bp,365.0bp)  .. (37);
  \draw [] (37) ..controls (128.5bp,382.74bp) and (147.26bp,388.02bp)  .. (38);
  \draw [] (38) ..controls (210.11bp,405.09bp) and (229.24bp,410.11bp)  .. (39);
  \draw [] (39) ..controls (291.94bp,425.99bp) and (311.33bp,430.72bp)  .. (40);
  \draw [] (40) ..controls (374.74bp,445.63bp) and (394.37bp,450.08bp)  .. (41);
  \draw [] (41) ..controls (458.06bp,463.88bp) and (477.43bp,467.91bp)  .. (42);
  \draw [] (42) ..controls (540.98bp,480.49bp) and (560.14bp,484.1bp)  .. (43);
  \draw [] (43) ..controls (623.65bp,495.64bp) and (642.18bp,498.87bp)  .. (44);
\end{tikzpicture}

%

%% file: figures/true_Lobsters.tex
%

\begin{tikzpicture}[anchor=mid,>=latex',]
\pgfsetlinewidth{1bp}
\node (0) at (127.77bp,167.36bp) [draw=graphnodecolor,fill=graphnodecolor,circle] {};
  \node (1) at (198.08bp,185.98bp) [draw=graphnodecolor,fill=graphnodecolor,circle] {};
  \node (2) at (193.88bp,295.75bp) [draw=graphnodecolor,fill=graphnodecolor,circle] {};
  \node (3) at (277.14bp,146.61bp) [draw=graphnodecolor,fill=graphnodecolor,circle] {};
  \node (5) at (171.09bp,96.15bp) [draw=graphnodecolor,fill=graphnodecolor,circle] {};
  \node (11) at (149.99bp,265.0bp) [draw=graphnodecolor,fill=graphnodecolor,circle] {};
  \node (12) at (224.31bp,253.92bp) [draw=graphnodecolor,fill=graphnodecolor,circle] {};
  \node (13) at (102.84bp,340.28bp) [draw=graphnodecolor,fill=graphnodecolor,circle] {};
  \node (21) at (281.4bp,347.5bp) [draw=graphnodecolor,fill=graphnodecolor,circle] {};
  \node (4) at (351.43bp,99.52bp) [draw=graphnodecolor,fill=graphnodecolor,circle] {};
  \node (6) at (81.805bp,81.65bp) [draw=graphnodecolor,fill=graphnodecolor,circle] {};
  \node (7) at (244.59bp,49.829bp) [draw=graphnodecolor,fill=graphnodecolor,circle] {};
  \node (8) at (106.31bp,41.532bp) [draw=graphnodecolor,fill=graphnodecolor,circle] {};
  \node (9) at (207.53bp,20.648bp) [draw=graphnodecolor,fill=graphnodecolor,circle] {};
  \node (10) at (154.01bp,18.0bp) [draw=graphnodecolor,fill=graphnodecolor,circle] {};
  \node (14) at (97.032bp,433.71bp) [draw=graphnodecolor,fill=graphnodecolor,circle] {};
  \node (15) at (54.674bp,280.68bp) [draw=graphnodecolor,fill=graphnodecolor,circle] {};
  \node (16) at (123.98bp,406.34bp) [draw=graphnodecolor,fill=graphnodecolor,circle] {};
  \node (17) at (57.359bp,413.54bp) [draw=graphnodecolor,fill=graphnodecolor,circle] {};
  \node (18) at (18.0bp,342.66bp) [draw=graphnodecolor,fill=graphnodecolor,circle] {};
  \node (19) at (30.126bp,382.03bp) [draw=graphnodecolor,fill=graphnodecolor,circle] {};
  \node (20) at (21.475bp,301.31bp) [draw=graphnodecolor,fill=graphnodecolor,circle] {};
  \node (22) at (350.55bp,385.99bp) [draw=graphnodecolor,fill=graphnodecolor,circle] {};
  \node (23) at (333.18bp,419.58bp) [draw=graphnodecolor,fill=graphnodecolor,circle] {};
  \node (24) at (258.37bp,434.48bp) [draw=graphnodecolor,fill=graphnodecolor,circle] {};
  \node (25) at (343.65bp,283.31bp) [draw=graphnodecolor,fill=graphnodecolor,circle] {};
  \node (26) at (371.16bp,354.55bp) [draw=graphnodecolor,fill=graphnodecolor,circle] {};
  \node (27) at (240.12bp,398.25bp) [draw=graphnodecolor,fill=graphnodecolor,circle] {};
  \node (28) at (305.47bp,294.78bp) [draw=graphnodecolor,fill=graphnodecolor,circle] {};
  \node (29) at (366.73bp,315.71bp) [draw=graphnodecolor,fill=graphnodecolor,circle] {};
  \node (30) at (296.35bp,431.14bp) [draw=graphnodecolor,fill=graphnodecolor,circle] {};
  \draw [] (0) ..controls (156.25bp,174.91bp) and (169.89bp,178.52bp)  .. (1);
  \draw [] (1) ..controls (196.61bp,224.45bp) and (195.35bp,257.25bp)  .. (2);
  \draw [] (1) ..controls (228.03bp,171.07bp) and (247.21bp,161.51bp)  .. (3);
  \draw [] (1) ..controls (188.1bp,152.76bp) and (181.06bp,129.31bp)  .. (5);
  \draw [] (2) ..controls (174.44bp,282.14bp) and (169.41bp,278.61bp)  .. (11);
  \draw [] (2) ..controls (207.47bp,277.07bp) and (210.67bp,272.67bp)  .. (12);
  \draw [] (2) ..controls (161.13bp,311.77bp) and (135.81bp,324.15bp)  .. (13);
  \draw [] (2) ..controls (225.45bp,314.42bp) and (250.01bp,328.94bp)  .. (21);
  \draw [] (3) ..controls (305.28bp,128.77bp) and (323.31bp,117.34bp)  .. (4);
  \draw [] (5) ..controls (137.53bp,90.7bp) and (115.33bp,87.094bp)  .. (6);
  \draw [] (5) ..controls (199.24bp,78.413bp) and (216.79bp,67.347bp)  .. (7);
  \draw [] (5) ..controls (146.17bp,75.138bp) and (131.2bp,62.516bp)  .. (8);
  \draw [] (5) ..controls (185.11bp,67.103bp) and (193.53bp,49.655bp)  .. (9);
  \draw [] (5) ..controls (164.42bp,65.627bp) and (160.66bp,48.437bp)  .. (10);
  \draw [] (13) ..controls (100.7bp,374.73bp) and (99.197bp,398.91bp)  .. (14);
  \draw [] (13) ..controls (83.75bp,316.65bp) and (73.669bp,304.18bp)  .. (15);
  \draw [] (13) ..controls (111.48bp,367.27bp) and (115.3bp,379.22bp)  .. (16);
  \draw [] (13) ..controls (85.479bp,368.25bp) and (74.707bp,385.6bp)  .. (17);
  \draw [] (13) ..controls (70.55bp,341.19bp) and (50.63bp,341.74bp)  .. (18);
  \draw [] (13) ..controls (74.868bp,356.34bp) and (58.064bp,365.99bp)  .. (19);
  \draw [] (13) ..controls (72.261bp,325.63bp) and (52.027bp,315.94bp)  .. (20);
  \draw [] (21) ..controls (308.61bp,362.65bp) and (323.45bp,370.91bp)  .. (22);
  \draw [] (21) ..controls (301.02bp,374.81bp) and (313.58bp,392.3bp)  .. (23);
  \draw [] (21) ..controls (272.74bp,380.2bp) and (267.02bp,401.82bp)  .. (24);
  \draw [] (21) ..controls (304.8bp,323.37bp) and (320.28bp,307.41bp)  .. (25);
  \draw [] (21) ..controls (315.14bp,350.15bp) and (337.46bp,351.9bp)  .. (26);
  \draw [] (21) ..controls (264.18bp,368.68bp) and (257.27bp,377.16bp)  .. (27);
  \draw [] (21) ..controls (291.76bp,324.82bp) and (295.03bp,317.65bp)  .. (28);
  \draw [] (21) ..controls (313.57bp,335.52bp) and (334.97bp,327.54bp)  .. (29);
  \draw [] (21) ..controls (287.09bp,379.34bp) and (290.6bp,398.97bp)  .. (30);
\end{tikzpicture}

%

%% file: figures/fake_Grids.tex
%

\begin{tikzpicture}[anchor=mid,>=latex',]
\pgfsetlinewidth{1bp}
\node (0) at (293.04bp,31.58bp) [draw=graphnodecolor,fill=graphnodecolor,circle] {};
  \node (1) at (205.16bp,22.127bp) [draw=graphnodecolor,fill=graphnodecolor,circle] {};
  \node (6) at (334.73bp,108.78bp) [draw=graphnodecolor,fill=graphnodecolor,circle] {};
  \node (2) at (116.89bp,18.0bp) [draw=graphnodecolor,fill=graphnodecolor,circle] {};
  \node (7) at (183.9bp,100.48bp) [draw=graphnodecolor,fill=graphnodecolor,circle] {};
  \node (3) at (32.111bp,21.07bp) [draw=graphnodecolor,fill=graphnodecolor,circle] {};
  \node (8) at (100.4bp,100.11bp) [draw=graphnodecolor,fill=graphnodecolor,circle] {};
  \node (9) at (18.0bp,105.03bp) [draw=graphnodecolor,fill=graphnodecolor,circle] {};
  \node (4) at (494.65bp,157.96bp) [draw=graphnodecolor,fill=graphnodecolor,circle] {};
  \node (5) at (415.98bp,132.96bp) [draw=graphnodecolor,fill=graphnodecolor,circle] {};
  \node (10) at (468.39bp,236.47bp) [draw=graphnodecolor,fill=graphnodecolor,circle] {};
  \node (11) at (386.14bp,211.98bp) [draw=graphnodecolor,fill=graphnodecolor,circle] {};
  \node (12) at (299.95bp,189.45bp) [draw=graphnodecolor,fill=graphnodecolor,circle] {};
  \node (13) at (209.47bp,181.14bp) [draw=graphnodecolor,fill=graphnodecolor,circle] {};
  \node (14) at (119.08bp,181.94bp) [draw=graphnodecolor,fill=graphnodecolor,circle] {};
  \node (15) at (33.025bp,188.4bp) [draw=graphnodecolor,fill=graphnodecolor,circle] {};
  \node (16) at (455.99bp,321.61bp) [draw=graphnodecolor,fill=graphnodecolor,circle] {};
  \node (17) at (375.58bp,296.5bp) [draw=graphnodecolor,fill=graphnodecolor,circle] {};
  \node (18) at (290.1bp,275.77bp) [draw=graphnodecolor,fill=graphnodecolor,circle] {};
  \node (19) at (201.85bp,265.9bp) [draw=graphnodecolor,fill=graphnodecolor,circle] {};
  \node (20) at (113.54bp,264.99bp) [draw=graphnodecolor,fill=graphnodecolor,circle] {};
  \node (21) at (30.672bp,273.86bp) [draw=graphnodecolor,fill=graphnodecolor,circle] {};
  \node (22) at (439.69bp,408.97bp) [draw=graphnodecolor,fill=graphnodecolor,circle] {};
  \node (23) at (364.01bp,381.1bp) [draw=graphnodecolor,fill=graphnodecolor,circle] {};
  \node (24) at (279.39bp,361.29bp) [draw=graphnodecolor,fill=graphnodecolor,circle] {};
  \node (25) at (192.22bp,350.35bp) [draw=graphnodecolor,fill=graphnodecolor,circle] {};
  \node (26) at (105.91bp,348.28bp) [draw=graphnodecolor,fill=graphnodecolor,circle] {};
  \node (27) at (30.332bp,362.63bp) [draw=graphnodecolor,fill=graphnodecolor,circle] {};
  \node (28) at (410.72bp,495.95bp) [draw=graphnodecolor,fill=graphnodecolor,circle] {};
  \node (29) at (347.5bp,464.61bp) [draw=graphnodecolor,fill=graphnodecolor,circle] {};
  \node (30) at (265.61bp,446.26bp) [draw=graphnodecolor,fill=graphnodecolor,circle] {};
  \node (31) at (181.11bp,434.98bp) [draw=graphnodecolor,fill=graphnodecolor,circle] {};
  \node (32) at (99.832bp,431.68bp) [draw=graphnodecolor,fill=graphnodecolor,circle] {};
  \node (33) at (46.871bp,455.33bp) [draw=graphnodecolor,fill=graphnodecolor,circle] {};
  \node (34) at (350.89bp,571.11bp) [draw=graphnodecolor,fill=graphnodecolor,circle] {};
  \node (35) at (318.5bp,544.64bp) [draw=graphnodecolor,fill=graphnodecolor,circle] {};
  \node (36) at (244.7bp,532.48bp) [draw=graphnodecolor,fill=graphnodecolor,circle] {};
  \node (37) at (164.58bp,522.05bp) [draw=graphnodecolor,fill=graphnodecolor,circle] {};
  \node (38) at (81.786bp,517.78bp) [draw=graphnodecolor,fill=graphnodecolor,circle] {};
  \node (39) at (123.54bp,540.99bp) [draw=graphnodecolor,fill=graphnodecolor,circle] {};
  \node (40) at (232.65bp,577.86bp) [draw=graphnodecolor,fill=graphnodecolor,circle] {};
  \node (41) at (280.53bp,623.45bp) [draw=graphnodecolor,fill=graphnodecolor,circle] {};
  \node (42) at (213.23bp,629.09bp) [draw=graphnodecolor,fill=graphnodecolor,circle] {};
  \node (43) at (133.81bp,615.17bp) [draw=graphnodecolor,fill=graphnodecolor,circle] {};
  \node (44) at (53.418bp,599.43bp) [draw=graphnodecolor,fill=graphnodecolor,circle] {};
  \draw [] (0) ..controls (259.75bp,27.999bp) and (238.42bp,25.705bp)  .. (1);
  \draw [] (0) ..controls (308.83bp,60.826bp) and (318.95bp,79.562bp)  .. (6);
  \draw [] (1) ..controls (171.72bp,20.563bp) and (150.3bp,19.562bp)  .. (2);
  \draw [] (1) ..controls (196.94bp,52.41bp) and (192.19bp,69.922bp)  .. (7);
  \draw [] (2) ..controls (84.179bp,19.184bp) and (64.415bp,19.9bp)  .. (3);
  \draw [] (2) ..controls (110.55bp,49.59bp) and (106.74bp,68.566bp)  .. (8);
  \draw [] (3) ..controls (26.724bp,53.122bp) and (23.382bp,73.006bp)  .. (9);
  \draw [] (4) ..controls (464.15bp,148.27bp) and (446.41bp,142.63bp)  .. (5);
  \draw [] (4) ..controls (484.44bp,188.48bp) and (478.47bp,206.34bp)  .. (10);
  \draw [] (5) ..controls (384.63bp,123.63bp) and (365.69bp,117.99bp)  .. (6);
  \draw [] (5) ..controls (404.46bp,163.45bp) and (397.51bp,181.87bp)  .. (11);
  \draw [] (6) ..controls (321.55bp,139.34bp) and (313.11bp,158.92bp)  .. (12);
  \draw [] (7) ..controls (151.78bp,100.34bp) and (132.48bp,100.25bp)  .. (8);
  \draw [] (7) ..controls (193.77bp,131.6bp) and (199.73bp,150.4bp)  .. (13);
  \draw [] (8) ..controls (68.457bp,102.02bp) and (49.879bp,103.13bp)  .. (9);
  \draw [] (8) ..controls (107.59bp,131.59bp) and (111.9bp,150.5bp)  .. (14);
  \draw [] (9) ..controls (23.798bp,137.2bp) and (27.3bp,156.63bp)  .. (15);
  \draw [] (10) ..controls (436.99bp,227.12bp) and (417.51bp,221.32bp)  .. (11);
  \draw [] (10) ..controls (463.65bp,268.98bp) and (460.72bp,289.14bp)  .. (16);
  \draw [] (11) ..controls (353.74bp,203.51bp) and (332.31bp,197.91bp)  .. (12);
  \draw [] (11) ..controls (382.11bp,244.25bp) and (379.61bp,264.26bp)  .. (17);
  \draw [] (12) ..controls (265.84bp,186.32bp) and (243.15bp,184.23bp)  .. (13);
  \draw [] (12) ..controls (296.23bp,222.05bp) and (293.86bp,242.82bp)  .. (18);
  \draw [] (13) ..controls (175.5bp,181.44bp) and (153.02bp,181.64bp)  .. (14);
  \draw [] (13) ..controls (206.56bp,213.5bp) and (204.76bp,233.57bp)  .. (19);
  \draw [] (14) ..controls (86.225bp,184.41bp) and (65.845bp,185.94bp)  .. (15);
  \draw [] (14) ..controls (116.95bp,213.89bp) and (115.67bp,233.09bp)  .. (20);
  \draw [] (15) ..controls (32.126bp,221.02bp) and (31.569bp,241.26bp)  .. (21);
  \draw [] (16) ..controls (425.05bp,311.95bp) and (406.47bp,306.15bp)  .. (17);
  \draw [] (16) ..controls (449.79bp,354.8bp) and (445.8bp,376.19bp)  .. (22);
  \draw [] (17) ..controls (343.2bp,288.65bp) and (322.46bp,283.62bp)  .. (18);
  \draw [] (17) ..controls (371.16bp,328.8bp) and (368.42bp,348.83bp)  .. (23);
  \draw [] (18) ..controls (256.57bp,272.02bp) and (234.97bp,269.6bp)  .. (19);
  \draw [] (18) ..controls (286.01bp,308.42bp) and (283.47bp,328.67bp)  .. (24);
  \draw [] (19) ..controls (168.39bp,265.56bp) and (146.96bp,265.34bp)  .. (20);
  \draw [] (19) ..controls (198.18bp,298.04bp) and (195.92bp,317.87bp)  .. (25);
  \draw [] (20) ..controls (81.657bp,268.4bp) and (62.507bp,270.45bp)  .. (21);
  \draw [] (20) ..controls (110.6bp,297.04bp) and (108.84bp,316.29bp)  .. (26);
  \draw [] (21) ..controls (30.543bp,307.59bp) and (30.46bp,329.32bp)  .. (27);
  \draw [] (22) ..controls (410.13bp,398.09bp) and (393.48bp,391.95bp)  .. (23);
  \draw [] (22) ..controls (428.89bp,441.41bp) and (421.51bp,463.57bp)  .. (28);
  \draw [] (23) ..controls (332.05bp,373.62bp) and (311.69bp,368.85bp)  .. (24);
  \draw [] (23) ..controls (357.7bp,412.98bp) and (353.79bp,432.76bp)  .. (29);
  \draw [] (24) ..controls (246.37bp,357.14bp) and (225.21bp,354.49bp)  .. (25);
  \draw [] (24) ..controls (274.13bp,393.73bp) and (270.86bp,413.85bp)  .. (30);
  \draw [] (25) ..controls (159.27bp,349.56bp) and (138.83bp,349.07bp)  .. (26);
  \draw [] (25) ..controls (187.98bp,382.66bp) and (185.35bp,402.7bp)  .. (31);
  \draw [] (26) ..controls (75.949bp,353.97bp) and (60.134bp,356.97bp)  .. (27);
  \draw [] (26) ..controls (103.57bp,380.37bp) and (102.17bp,399.64bp)  .. (32);
  \draw [] (27) ..controls (36.45bp,396.92bp) and (40.766bp,421.11bp)  .. (33);
  \draw [] (28) ..controls (385.07bp,483.24bp) and (373.34bp,477.42bp)  .. (29);
  \draw [] (28) ..controls (388.84bp,523.44bp) and (372.97bp,543.37bp)  .. (34);
  \draw [] (29) ..controls (315.99bp,457.55bp) and (297.07bp,453.31bp)  .. (30);
  \draw [] (29) ..controls (336.43bp,495.16bp) and (329.56bp,514.11bp)  .. (35);
  \draw [] (30) ..controls (233.35bp,441.95bp) and (213.34bp,439.28bp)  .. (31);
  \draw [] (30) ..controls (257.66bp,479.02bp) and (252.54bp,500.13bp)  .. (36);
  \draw [] (31) ..controls (149.6bp,433.7bp) and (131.28bp,432.95bp)  .. (32);
  \draw [] (31) ..controls (174.83bp,468.06bp) and (170.78bp,489.38bp)  .. (37);
  \draw [] (32) ..controls (77.049bp,441.85bp) and (69.842bp,445.07bp)  .. (33);
  \draw [] (32) ..controls (92.996bp,464.29bp) and (88.617bp,485.19bp)  .. (38);
  \draw [] (33) ..controls (73.581bp,485.17bp) and (97.236bp,511.6bp)  .. (39);
  \draw [] (34) ..controls (335.23bp,558.32bp) and (333.86bp,557.19bp)  .. (35);
  \draw [] (34) ..controls (310.39bp,573.42bp) and (272.9bp,575.56bp)  .. (40);
  \draw [] (35) ..controls (289.12bp,539.8bp) and (274.19bp,537.34bp)  .. (36);
  \draw [] (35) ..controls (304.12bp,574.49bp) and (294.9bp,593.62bp)  .. (41);
  \draw [] (36) ..controls (213.41bp,528.41bp) and (195.78bp,526.11bp)  .. (37);
  \draw [] (36) ..controls (233.34bp,567.34bp) and (224.52bp,594.44bp)  .. (42);
  \draw [] (37) ..controls (132.39bp,520.39bp) and (113.56bp,519.42bp)  .. (38);
  \draw [] (37) ..controls (153.25bp,556.32bp) and (144.96bp,581.43bp)  .. (43);
  \draw [] (38) ..controls (101.03bp,528.47bp) and (104.41bp,530.35bp)  .. (39);
  \draw [] (38) ..controls (70.924bp,549.04bp) and (64.147bp,568.55bp)  .. (44);
  \draw [] (39) ..controls (161.55bp,553.83bp) and (195.22bp,565.21bp)  .. (40);
  \draw [] (40) ..controls (252.49bp,596.75bp) and (260.71bp,604.58bp)  .. (41);
  \draw [] (41) ..controls (252.84bp,625.77bp) and (240.98bp,626.77bp)  .. (42);
  \draw [] (42) ..controls (182.21bp,623.66bp) and (164.74bp,620.59bp)  .. (43);
  \draw [] (43) ..controls (102.64bp,609.07bp) and (84.519bp,605.52bp)  .. (44);
\end{tikzpicture}

%

%% file: figures/fake_Grids_with_info.tex
%

\begin{tikzpicture}[anchor=mid,>=latex',]
\pgfsetlinewidth{1bp}
\node (0) at (98.064bp,18.0bp) [draw=graphnodecolor,fill=graphnodecolor,circle] {};
  \node (1) at (178.77bp,34.179bp) [draw=graphnodecolor,fill=graphnodecolor,circle] {};
  \node (9) at (79.006bp,100.42bp) [draw=graphnodecolor,fill=graphnodecolor,circle] {};
  \node (2) at (260.83bp,51.421bp) [draw=graphnodecolor,fill=graphnodecolor,circle] {};
  \node (10) at (160.1bp,118.74bp) [draw=graphnodecolor,fill=graphnodecolor,circle] {};
  \node (3) at (343.06bp,69.72bp) [draw=graphnodecolor,fill=graphnodecolor,circle] {};
  \node (11) at (241.97bp,137.18bp) [draw=graphnodecolor,fill=graphnodecolor,circle] {};
  \node (4) at (425.19bp,88.733bp) [draw=graphnodecolor,fill=graphnodecolor,circle] {};
  \node (12) at (323.55bp,156.05bp) [draw=graphnodecolor,fill=graphnodecolor,circle] {};
  \node (5) at (507.18bp,108.32bp) [draw=graphnodecolor,fill=graphnodecolor,circle] {};
  \node (13) at (404.91bp,175.11bp) [draw=graphnodecolor,fill=graphnodecolor,circle] {};
  \node (6) at (588.93bp,128.52bp) [draw=graphnodecolor,fill=graphnodecolor,circle] {};
  \node (14) at (486.23bp,194.33bp) [draw=graphnodecolor,fill=graphnodecolor,circle] {};
  \node (7) at (670.08bp,149.53bp) [draw=graphnodecolor,fill=graphnodecolor,circle] {};
  \node (15) at (567.68bp,213.74bp) [draw=graphnodecolor,fill=graphnodecolor,circle] {};
  \node (8) at (749.56bp,170.82bp) [draw=graphnodecolor,fill=graphnodecolor,circle] {};
  \node (16) at (649.21bp,233.58bp) [draw=graphnodecolor,fill=graphnodecolor,circle] {};
  \node (17) at (730.0bp,253.13bp) [draw=graphnodecolor,fill=graphnodecolor,circle] {};
  \node (18) at (58.756bp,184.57bp) [draw=graphnodecolor,fill=graphnodecolor,circle] {};
  \node (19) at (139.69bp,204.57bp) [draw=graphnodecolor,fill=graphnodecolor,circle] {};
  \node (20) at (221.39bp,224.67bp) [draw=graphnodecolor,fill=graphnodecolor,circle] {};
  \node (21) at (302.82bp,244.48bp) [draw=graphnodecolor,fill=graphnodecolor,circle] {};
  \node (22) at (384.08bp,263.84bp) [draw=graphnodecolor,fill=graphnodecolor,circle] {};
  \node (23) at (465.46bp,282.75bp) [draw=graphnodecolor,fill=graphnodecolor,circle] {};
  \node (24) at (547.18bp,301.25bp) [draw=graphnodecolor,fill=graphnodecolor,circle] {};
  \node (25) at (629.31bp,319.53bp) [draw=graphnodecolor,fill=graphnodecolor,circle] {};
  \node (26) at (710.72bp,337.52bp) [draw=graphnodecolor,fill=graphnodecolor,circle] {};
  \node (27) at (38.664bp,268.7bp) [draw=graphnodecolor,fill=graphnodecolor,circle] {};
  \node (28) at (119.12bp,290.33bp) [draw=graphnodecolor,fill=graphnodecolor,circle] {};
  \node (29) at (200.41bp,312.0bp) [draw=graphnodecolor,fill=graphnodecolor,circle] {};
  \node (30) at (281.79bp,332.75bp) [draw=graphnodecolor,fill=graphnodecolor,circle] {};
  \node (31) at (363.28bp,352.46bp) [draw=graphnodecolor,fill=graphnodecolor,circle] {};
  \node (32) at (445.01bp,371.15bp) [draw=graphnodecolor,fill=graphnodecolor,circle] {};
  \node (33) at (527.12bp,388.8bp) [draw=graphnodecolor,fill=graphnodecolor,circle] {};
  \node (34) at (609.58bp,405.49bp) [draw=graphnodecolor,fill=graphnodecolor,circle] {};
  \node (35) at (691.31bp,421.8bp) [draw=graphnodecolor,fill=graphnodecolor,circle] {};
  \node (36) at (18.0bp,350.59bp) [draw=graphnodecolor,fill=graphnodecolor,circle] {};
  \node (37) at (97.272bp,373.97bp) [draw=graphnodecolor,fill=graphnodecolor,circle] {};
  \node (38) at (178.45bp,396.78bp) [draw=graphnodecolor,fill=graphnodecolor,circle] {};
  \node (39) at (260.51bp,418.32bp) [draw=graphnodecolor,fill=graphnodecolor,circle] {};
  \node (40) at (343.08bp,438.47bp) [draw=graphnodecolor,fill=graphnodecolor,circle] {};
  \node (41) at (425.99bp,457.23bp) [draw=graphnodecolor,fill=graphnodecolor,circle] {};
  \node (42) at (509.09bp,474.48bp) [draw=graphnodecolor,fill=graphnodecolor,circle] {};
  \node (43) at (591.99bp,490.11bp) [draw=graphnodecolor,fill=graphnodecolor,circle] {};
  \node (44) at (673.43bp,504.33bp) [draw=graphnodecolor,fill=graphnodecolor,circle] {};
  \draw [] (0) ..controls (129.35bp,24.272bp) and (147.55bp,27.92bp)  .. (1);
  \draw [] (0) ..controls (90.71bp,49.803bp) and (86.267bp,69.016bp)  .. (9);
  \draw [] (1) ..controls (210.34bp,40.813bp) and (229.3bp,44.797bp)  .. (2);
  \draw [] (1) ..controls (171.62bp,66.558bp) and (167.16bp,86.758bp)  .. (10);
  \draw [] (2) ..controls (292.46bp,58.461bp) and (311.46bp,62.69bp)  .. (3);
  \draw [] (2) ..controls (253.68bp,83.91bp) and (249.11bp,104.72bp)  .. (11);
  \draw [] (3) ..controls (374.65bp,77.035bp) and (393.63bp,81.428bp)  .. (4);
  \draw [] (3) ..controls (335.64bp,102.52bp) and (330.87bp,123.65bp)  .. (12);
  \draw [] (4) ..controls (456.73bp,96.267bp) and (475.68bp,100.79bp)  .. (5);
  \draw [] (4) ..controls (417.48bp,121.55bp) and (412.52bp,142.7bp)  .. (13);
  \draw [] (5) ..controls (538.63bp,116.09bp) and (557.52bp,120.76bp)  .. (6);
  \draw [] (5) ..controls (499.22bp,141.0bp) and (494.09bp,162.05bp)  .. (14);
  \draw [] (6) ..controls (620.15bp,136.6bp) and (638.9bp,141.46bp)  .. (7);
  \draw [] (6) ..controls (580.88bp,160.8bp) and (575.72bp,181.49bp)  .. (15);
  \draw [] (7) ..controls (700.89bp,157.78bp) and (718.81bp,162.58bp)  .. (8);
  \draw [] (7) ..controls (662.09bp,181.71bp) and (657.1bp,201.79bp)  .. (16);
  \draw [] (8) ..controls (742.01bp,202.58bp) and (737.45bp,221.77bp)  .. (17);
  \draw [] (9) ..controls (110.2bp,107.47bp) and (128.94bp,111.7bp)  .. (10);
  \draw [] (9) ..controls (71.252bp,132.64bp) and (66.414bp,152.75bp)  .. (18);
  \draw [] (10) ..controls (191.6bp,125.83bp) and (210.52bp,130.1bp)  .. (11);
  \draw [] (10) ..controls (152.37bp,151.25bp) and (147.41bp,172.08bp)  .. (19);
  \draw [] (11) ..controls (273.36bp,144.44bp) and (292.21bp,148.8bp)  .. (12);
  \draw [] (11) ..controls (234.24bp,170.07bp) and (229.12bp,191.82bp)  .. (20);
  \draw [] (12) ..controls (354.85bp,163.38bp) and (373.65bp,167.79bp)  .. (13);
  \draw [] (12) ..controls (315.73bp,189.39bp) and (310.53bp,211.56bp)  .. (21);
  \draw [] (13) ..controls (436.2bp,182.5bp) and (454.99bp,186.94bp)  .. (14);
  \draw [] (13) ..controls (397.17bp,208.09bp) and (391.91bp,230.51bp)  .. (22);
  \draw [] (14) ..controls (517.56bp,201.8bp) and (536.38bp,206.28bp)  .. (15);
  \draw [] (14) ..controls (478.4bp,227.66bp) and (473.19bp,249.84bp)  .. (23);
  \draw [] (15) ..controls (599.05bp,221.37bp) and (617.89bp,225.96bp)  .. (16);
  \draw [] (15) ..controls (559.97bp,246.63bp) and (554.88bp,268.39bp)  .. (24);
  \draw [] (16) ..controls (680.3bp,241.1bp) and (698.96bp,245.62bp)  .. (17);
  \draw [] (16) ..controls (641.67bp,266.14bp) and (636.84bp,287.0bp)  .. (25);
  \draw [] (17) ..controls (722.62bp,285.44bp) and (718.01bp,305.6bp)  .. (26);
  \draw [] (18) ..controls (89.893bp,192.27bp) and (108.6bp,196.89bp)  .. (19);
  \draw [] (18) ..controls (51.062bp,216.79bp) and (46.263bp,236.88bp)  .. (27);
  \draw [] (19) ..controls (171.12bp,212.31bp) and (190.0bp,216.95bp)  .. (20);
  \draw [] (19) ..controls (131.9bp,237.06bp) and (126.91bp,257.87bp)  .. (28);
  \draw [] (20) ..controls (252.72bp,232.29bp) and (271.53bp,236.87bp)  .. (21);
  \draw [] (20) ..controls (213.51bp,257.5bp) and (208.29bp,279.21bp)  .. (29);
  \draw [] (21) ..controls (334.08bp,251.93bp) and (352.86bp,256.4bp)  .. (22);
  \draw [] (21) ..controls (294.89bp,277.76bp) and (289.61bp,299.9bp)  .. (30);
  \draw [] (22) ..controls (415.39bp,271.11bp) and (434.19bp,275.48bp)  .. (23);
  \draw [] (22) ..controls (376.24bp,297.25bp) and (371.02bp,319.48bp)  .. (31);
  \draw [] (23) ..controls (496.9bp,289.87bp) and (515.78bp,294.14bp)  .. (24);
  \draw [] (23) ..controls (457.75bp,316.07bp) and (452.62bp,338.25bp)  .. (32);
  \draw [] (24) ..controls (578.78bp,308.28bp) and (597.76bp,312.51bp)  .. (25);
  \draw [] (24) ..controls (539.64bp,334.16bp) and (534.66bp,355.93bp)  .. (33);
  \draw [] (25) ..controls (660.63bp,326.45bp) and (679.44bp,330.61bp)  .. (26);
  \draw [] (25) ..controls (621.84bp,352.09bp) and (617.05bp,372.96bp)  .. (34);
  \draw [] (26) ..controls (703.29bp,369.79bp) and (698.65bp,389.92bp)  .. (35);
  \draw [] (27) ..controls (69.618bp,277.02bp) and (88.211bp,282.02bp)  .. (28);
  \draw [] (27) ..controls (30.714bp,300.21bp) and (25.939bp,319.13bp)  .. (36);
  \draw [] (28) ..controls (150.4bp,298.67bp) and (169.18bp,303.68bp)  .. (29);
  \draw [] (28) ..controls (110.76bp,322.36bp) and (105.54bp,342.34bp)  .. (37);
  \draw [] (29) ..controls (231.72bp,319.99bp) and (250.52bp,324.78bp)  .. (30);
  \draw [] (29) ..controls (192.09bp,344.12bp) and (186.76bp,364.7bp)  .. (38);
  \draw [] (30) ..controls (313.14bp,340.33bp) and (331.97bp,344.89bp)  .. (31);
  \draw [] (30) ..controls (273.73bp,365.17bp) and (268.56bp,385.93bp)  .. (39);
  \draw [] (31) ..controls (394.72bp,359.65bp) and (413.61bp,363.97bp)  .. (32);
  \draw [] (31) ..controls (355.63bp,385.04bp) and (350.73bp,405.92bp)  .. (40);
  \draw [] (32) ..controls (476.6bp,377.94bp) and (495.58bp,382.02bp)  .. (33);
  \draw [] (32) ..controls (437.81bp,403.76bp) and (433.19bp,424.65bp)  .. (41);
  \draw [] (33) ..controls (558.85bp,395.22bp) and (577.9bp,399.08bp)  .. (34);
  \draw [] (33) ..controls (520.29bp,421.26bp) and (515.92bp,442.05bp)  .. (42);
  \draw [] (34) ..controls (641.02bp,411.76bp) and (659.91bp,415.53bp)  .. (35);
  \draw [] (34) ..controls (602.85bp,437.89bp) and (598.64bp,458.11bp)  .. (43);
  \draw [] (35) ..controls (684.43bp,453.55bp) and (680.3bp,472.62bp)  .. (44);
  \draw [] (36) ..controls (48.821bp,359.68bp) and (66.852bp,365.0bp)  .. (37);
  \draw [] (37) ..controls (128.5bp,382.74bp) and (147.26bp,388.02bp)  .. (38);
  \draw [] (38) ..controls (210.11bp,405.09bp) and (229.24bp,410.11bp)  .. (39);
  \draw [] (39) ..controls (291.94bp,425.99bp) and (311.33bp,430.72bp)  .. (40);
  \draw [] (40) ..controls (374.74bp,445.63bp) and (394.37bp,450.08bp)  .. (41);
  \draw [] (41) ..controls (458.06bp,463.88bp) and (477.43bp,467.91bp)  .. (42);
  \draw [] (42) ..controls (540.98bp,480.49bp) and (560.14bp,484.1bp)  .. (43);
  \draw [] (43) ..controls (623.65bp,495.64bp) and (642.18bp,498.87bp)  .. (44);
\end{tikzpicture}

%

%% file: figures/fake_Lobsters.tex
%

\begin{tikzpicture}[anchor=mid,>=latex',]
\pgfsetlinewidth{1bp}
\node (0) at (188.93bp,281.0bp) [draw=graphnodecolor,fill=graphnodecolor,circle] {};
  \node (1) at (263.56bp,341.62bp) [draw=graphnodecolor,fill=graphnodecolor,circle] {};
  \node (2) at (258.94bp,210.44bp) [draw=graphnodecolor,fill=graphnodecolor,circle] {};
  \node (3) at (154.13bp,189.93bp) [draw=graphnodecolor,fill=graphnodecolor,circle] {};
  \node (5) at (156.71bp,357.76bp) [draw=graphnodecolor,fill=graphnodecolor,circle] {};
  \node (6) at (109.4bp,263.26bp) [draw=graphnodecolor,fill=graphnodecolor,circle] {};
  \node (8) at (111.79bp,311.96bp) [draw=graphnodecolor,fill=graphnodecolor,circle] {};
  \node (25) at (209.11bp,241.94bp) [draw=graphnodecolor,fill=graphnodecolor,circle] {};
  \node (4) at (324.22bp,419.25bp) [draw=graphnodecolor,fill=graphnodecolor,circle] {};
  \node (11) at (237.71bp,407.25bp) [draw=graphnodecolor,fill=graphnodecolor,circle] {};
  \node (15) at (333.16bp,335.19bp) [draw=graphnodecolor,fill=graphnodecolor,circle] {};
  \node (28) at (330.65bp,296.4bp) [draw=graphnodecolor,fill=graphnodecolor,circle] {};
  \node (9) at (322.44bp,143.62bp) [draw=graphnodecolor,fill=graphnodecolor,circle] {};
  \node (21) at (258.64bp,129.43bp) [draw=graphnodecolor,fill=graphnodecolor,circle] {};
  \node (26) at (339.23bp,203.21bp) [draw=graphnodecolor,fill=graphnodecolor,circle] {};
  \node (7) at (167.91bp,99.874bp) [draw=graphnodecolor,fill=graphnodecolor,circle] {};
  \node (10) at (111.15bp,124.62bp) [draw=graphnodecolor,fill=graphnodecolor,circle] {};
  \node (17) at (71.761bp,158.36bp) [draw=graphnodecolor,fill=graphnodecolor,circle] {};
  \node (13) at (406.5bp,436.84bp) [draw=graphnodecolor,fill=graphnodecolor,circle] {};
  \node (14) at (326.7bp,503.44bp) [draw=graphnodecolor,fill=graphnodecolor,circle] {};
  \node (22) at (375.09bp,480.62bp) [draw=graphnodecolor,fill=graphnodecolor,circle] {};
  \node (27) at (280.15bp,498.08bp) [draw=graphnodecolor,fill=graphnodecolor,circle] {};
  \node (29) at (411.76bp,389.14bp) [draw=graphnodecolor,fill=graphnodecolor,circle] {};
  \node (12) at (139.46bp,443.17bp) [draw=graphnodecolor,fill=graphnodecolor,circle] {};
  \node (30) at (86.971bp,418.73bp) [draw=graphnodecolor,fill=graphnodecolor,circle] {};
  \node (20) at (29.524bp,220.42bp) [draw=graphnodecolor,fill=graphnodecolor,circle] {};
  \node (24) at (26.958bp,310.19bp) [draw=graphnodecolor,fill=graphnodecolor,circle] {};
  \node (16) at (151.09bp,18.0bp) [draw=graphnodecolor,fill=graphnodecolor,circle] {};
  \node (18) at (18.0bp,288.71bp) [draw=graphnodecolor,fill=graphnodecolor,circle] {};
  \node (19) at (43.888bp,374.66bp) [draw=graphnodecolor,fill=graphnodecolor,circle] {};
  \node (23) at (382.36bp,86.997bp) [draw=graphnodecolor,fill=graphnodecolor,circle] {};
  \draw [] (0) ..controls (216.31bp,303.24bp) and (236.26bp,319.45bp)  .. (1);
  \draw [] (0) ..controls (214.4bp,255.33bp) and (233.58bp,236.0bp)  .. (2);
  \draw [] (0) ..controls (176.16bp,247.59bp) and (166.86bp,223.24bp)  .. (3);
  \draw [] (0) ..controls (176.53bp,310.53bp) and (169.09bp,328.27bp)  .. (5);
  \draw [] (0) ..controls (158.1bp,274.12bp) and (140.17bp,270.12bp)  .. (6);
  \draw [] (0) ..controls (159.34bp,292.87bp) and (141.67bp,299.97bp)  .. (8);
  \draw [] (0) ..controls (198.49bp,262.49bp) and (199.65bp,260.26bp)  .. (25);
  \draw [] (1) ..controls (285.56bp,369.78bp) and (302.04bp,390.86bp)  .. (4);
  \draw [] (1) ..controls (253.07bp,368.24bp) and (248.27bp,380.43bp)  .. (11);
  \draw [] (1) ..controls (291.79bp,339.01bp) and (304.71bp,337.82bp)  .. (15);
  \draw [] (1) ..controls (289.76bp,323.95bp) and (304.52bp,314.01bp)  .. (28);
  \draw [] (2) ..controls (282.62bp,185.53bp) and (298.8bp,168.5bp)  .. (9);
  \draw [] (2) ..controls (258.82bp,178.71bp) and (258.76bp,160.73bp)  .. (21);
  \draw [] (2) ..controls (290.3bp,207.62bp) and (307.96bp,206.03bp)  .. (26);
  \draw [] (3) ..controls (159.32bp,155.98bp) and (162.78bp,133.39bp)  .. (7);
  \draw [] (3) ..controls (137.22bp,164.23bp) and (128.0bp,150.22bp)  .. (10);
  \draw [] (3) ..controls (122.93bp,177.97bp) and (102.94bp,170.31bp)  .. (17);
  \draw [] (4) ..controls (355.88bp,426.02bp) and (374.89bp,430.08bp)  .. (13);
  \draw [] (4) ..controls (325.18bp,451.64bp) and (325.75bp,471.09bp)  .. (14);
  \draw [] (4) ..controls (344.09bp,443.22bp) and (355.28bp,456.72bp)  .. (22);
  \draw [] (4) ..controls (307.66bp,448.88bp) and (296.7bp,468.48bp)  .. (27);
  \draw [] (4) ..controls (356.86bp,408.02bp) and (379.16bp,400.35bp)  .. (29);
  \draw [] (5) ..controls (150.18bp,390.12bp) and (145.99bp,410.85bp)  .. (12);
  \draw [] (5) ..controls (130.7bp,380.49bp) and (112.94bp,396.03bp)  .. (30);
  \draw [] (6) ..controls (79.285bp,247.11bp) and (59.252bp,236.36bp)  .. (20);
  \draw [] (6) ..controls (78.904bp,280.62bp) and (57.387bp,292.87bp)  .. (24);
  \draw [] (7) ..controls (161.44bp,68.375bp) and (157.55bp,49.455bp)  .. (16);
  \draw [] (8) ..controls (77.383bp,303.43bp) and (52.305bp,297.21bp)  .. (18);
  \draw [] (8) ..controls (86.473bp,335.34bp) and (69.172bp,351.31bp)  .. (19);
  \draw [] (9) ..controls (345.67bp,121.67bp) and (359.18bp,108.9bp)  .. (23);
\end{tikzpicture}

%